%% file: cvpr.tex
\documentclass[10pt,twocolumn,letterpaper]{article}

\usepackage[pagenumbers]{cvpr} 

\usepackage{graphicx}
\usepackage{amsmath}
\usepackage{amssymb}
\usepackage{booktabs}
\usepackage[misc]{ifsym}

\usepackage[pagebackref,breaklinks,colorlinks]{hyperref}

\makeatletter
\def\UrlAlphabet{%
      \do\a\do\b\do\c\do\d\do\e\do\f\do\g\do\h\do\i\do\j%
      \do\k\do\l\do\m\do\n\do\o\do\p\do\q\do\r\do\s\do\t%
      \do\u\do\v\do\w\do\x\do\y\do\z\do\A\do\B\do\C\do\D%
      \do\E\do\F\do\G\do\H\do\I\do\J\do\K\do\L\do\M\do\N%
      \do\O\do\P\do\Q\do\R\do\S\do\T\do\U\do\V\do\W\do\X%
      \do\Y\do\Z}
\def\UrlDigits{\do\1\do\2\do\3\do\4\do\5\do\6\do\7\do\8\do\9\do\0}
\g@addto@macro{\UrlBreaks}{\UrlOrds}
\g@addto@macro{\UrlBreaks}{\UrlAlphabet}
\g@addto@macro{\UrlBreaks}{\UrlDigits}
\makeatother

\usepackage[capitalize]{cleveref}
\crefname{section}{Sec.}{Secs.}
\Crefname{section}{Section}{Sections}
\Crefname{table}{Table}{Tables}
\crefname{table}{Tab.}{Tabs.}

\def\cvprPaperID{8865} 
\def\confName{CVPR}
\def\confYear{2022}

\usepackage{amsmath}
\usepackage{amssymb}
\usepackage{booktabs}

\input{tables.tex}
\input{figures.tex}

\begin{document}

\title{AdaMixer: A Fast-Converging Query-Based Object Detector}

\author{
Ziteng Gao\textsuperscript{1} \quad \quad Limin Wang\textsuperscript{1 \Letter} \quad \quad Bing Han\textsuperscript{2} \quad \quad Sheng Guo\textsuperscript{2}\\
\textsuperscript{1}State Key Laboratory for Novel Software Technology, Nanjing University, China \\
\textsuperscript{2}MYbank, Ant Group, China
}

\maketitle

\begin{abstract}
Traditional object detectors employ the dense paradigm of scanning over locations and scales in an image. The recent query-based object detectors break this convention by decoding image features with a set of learnable queries. However, this paradigm still suffers from slow convergence, limited performance, and design complexity of extra networks between backbone and decoder. In this paper, we find that the key to these issues is the adaptability of decoders for casting queries to varying objects. Accordingly, we propose a fast-converging query-based detector, named AdaMixer, by improving the adaptability of query-based decoding processes in two aspects. First, each query adaptively samples features over space and scales based on estimated offsets, which allows AdaMixer to efficiently attend to the coherent regions of objects. Then, we dynamically decode these sampled features with an adaptive MLP-Mixer under the guidance of each query. Thanks to these two critical designs, AdaMixer enjoys architectural simplicity without requiring dense attentional encoders or explicit pyramid networks. On the challenging MS COCO benchmark, AdaMixer with ResNet-50 as the backbone, with 12 training epochs, reaches up to 45.0 AP on the validation set along with 27.9 AP$_s$ in detecting small objects. With the longer training scheme, AdaMixer with ResNeXt-101-DCN and Swin-S reaches 49.5 and 51.3 AP. Our work sheds light on a simple, accurate, and fast converging architecture for query-based object detectors.
The code is made available at \url{https://github.com/MCG-NJU/AdaMixer}.
\end{abstract}
\blfootnote{ \Letter: Corresponding author (lmwang@nju.edu.cn).}

\section{Introduction}
\FigureCurve
Object detection has been a fundamental task in the computer vision area for decades, as it aims to locate varying objects in a single image and categorize them.
For a long time, researchers have used spatial dense prior on grids in an image to cover potential objects with great variations.
This dense paradigm dates back from sliding window methods~\cite{DBLP:journals/pami/FelzenszwalbGMR10, DBLP:conf/cvpr/ViolaJ01, DBLP:journals/corr/SermanetEZMFL13} and still remains prevalent in anchor-based or point-based detectors~\cite{DBLP:conf/iccv/Girshick15, DBLP:conf/nips/RenHGS15, DBLP:conf/iccv/LinGGHD17, DBLP:conf/eccv/LiuAESRFB16, DBLP:conf/cvpr/RedmonDGF16, DBLP:conf/iccv/TianSCH19, DBLP:journals/corr/abs-1904-07850} at the age of convolutional neural networks.
Although dense priors has been a dominant role in object detection for its remarkable performance to cover potential objects, they are criticized for several shortcomings in various aspects, including anchor designs~\cite{DBLP:conf/nips/YangZLZS18, DBLP:journals/corr/abs-1804-02767, DBLP:conf/cvpr/WangCYLL19}, training sample selection~\cite{DBLP:conf/nips/ZhangWLJY19, DBLP:journals/corr/abs-1912-02424, DBLP:conf/cvpr/GeLLYS21,DBLP:conf/iccv/Gao0W21}, and post-processing operators on potential redundant detections~\cite{DBLP:conf/iccv/BodlaSCD17, DBLP:conf/cvpr/HeZWS019}.

Though sorts of remedies on these issues are proposed every year, the underlying detection scheme with dense grid-like prior had remained almost untouched for a long time.
Recently, query-based object detectors~\cite{DBLP:conf/eccv/CarionMSUKZ20, DBLP:journals/corr/abs-2010-04159, DBLP:journals/corr/abs-2011-12450} bring a new perspective on object detection, that is, to use learnable embeddings, also termed queries, to directly represent potential objects by using attention-like operators~\cite{DBLP:conf/nips/VaswaniSPUJGKP17}.
This scheme, on the other hand, requires a strong representation power of the network to cast limited queries to potential objects to cope with great variations of objects across images.
However, the adaptability of currently employed query decoders to the image content is limited on both how to spatially sample features and how to process sampled features.
For instance, attention-based decoder in DETR-like detectors~\cite{DBLP:conf/eccv/CarionMSUKZ20, DBLP:journals/corr/abs-2010-04159} are adaptive on which feature to sample but remain static on how to decode it, whereas dynamic interaction head in Sparse R-CNN~\cite{DBLP:journals/corr/abs-2011-12450} goes vice versa.
The insufficiency in the adaptability to different images leaves the current decoder in a dilemma between limited query representation power and great variations of objects.
Also, as compensation for this, query-based object detectors usually bring extra attentional encoders or explicit pyramid necks after the backbone and before the query decoder, in order to involve more semantic or multi-scale modeling, such as TransformerEncoder~\cite{DBLP:conf/nips/VaswaniSPUJGKP17}, MultiScaleDeformableTransformerEncoder~\cite{DBLP:journals/corr/abs-2010-04159} and FPN~\cite{DBLP:conf/cvpr/LinDGHHB17}.
These extra components result in the higher complexity of built detection pipelines in both design and computational aspects. In addition, detectors with them are hungry for more training time and rich data augmentation due to the introduced modules.

In this paper, we present a fast-converging and accurate query-based object detector with a simplified architecture, named AdaMixer, to mitigate issues above.
Specifically, to effectively use queries to represent objects, AdaMixer introduces the adaptive 3D feature sampler and the adaptive mixing of channel semantics and spatial structures holistically.
First, by regarding multi-scale feature maps from the backbone as a 3D feature space, our proposed decoder can flexibly sample features over space and scales to adaptively handle both of location and scale variations of objects based on queries.
Then, the adaptive mixing applies the channel and spatial mixing to the sampled features with dynamic kernels under the guidance of queries.
The adaptive location sampling and holistic content decoding notably enhances the adaptability of queries to varying images in detecting varying objects.
As a result, AdaMixer is simply made up of a backbone network and our proposed decoder without extra attentional encoders or explicit pyramid networks.

Experimental results show that in a standard 12 epochs training ($1\times$ training scheme) with the random flipping as the only augmentation, our AdaMixer with ResNet-50~\cite{DBLP:conf/cvpr/HeZRS16} as the backbone achieves 42.7, 44.1, and 45.0 AP on MS COCO validation set under the settings of 100, 300, and 500 queries, with 24.7, 27.0, and 27.9 AP$_s$ in small object detection.
With longer $3\times$ training time and stronger data augmentation aligned to other query-based detectors, our AdaMixer with ResNet-101, ResNeXt-101-DCN~\cite{DBLP:conf/cvpr/XieGDTH17, DBLP:journals/corr/abs-1811-11168}, and Swin-S~\cite{DBLP:journals/corr/abs-2103-14030} achieves 48.0, 49.5, and 51.3 AP with the single scale and single model testing, significantly outperforming the previous state-of-the-art query-based detectors. 
We hope AdaMixer, as a simply-designed, fast-converging, relatively efficient, and more accurate object detector, will serve as a strong baseline in the future research for the query-based object detection.

\TableComparisonToOtherSparseDetectors

\input{related.tex}
\input{method.tex}
\input{experiment.tex}

\section{Conclusion and Limitation}
In this paper, we have presented a fast-converging query-based object detection architecture, termed AdaMixer, to efficiently and effectively decode objects from images.
Our proposed AdaMixer improves the decoder of query-based detectors with adaptive 3D sampling and adaptive channel and spatial mixing.
By improving query decoders, AdaMixer circumvents the requirement of extra network modeling between backbone and decoder. Our AdaMixer achieves superior performance, especially on small object detection, with less computational cost compared to other query-based detectors.
Moreover, it enables the fast convergence speed with limited training budgets. We hope that AdaMixer can serve as a strong baseline for future research. 

The limitation in our AdaMixer is that though we have applied the grouping mechanism, the total parameter number remains a little bit large. This is mainly due to a large number of parameters in the linear layer to produce dynamic mixing weights. We leave the question of how to further reduce the number of parameters to the future work.

\paragraph {\bf Acknowledgements.} {\small This work is supported by National Natural Science Foundation of China  (No.62076119, No.61921006),  Program for Innovative Talents and Entrepreneur in Jiangsu Province, and Collaborative Innovation Center of Novel Software Technology and Industrialization. Part of the work is done during the internship of Ziteng at MYbank.}

\input{supp.tex}
{\small
\bibliographystyle{ieee_fullname}
\bibliography{egbib}
}

\end{document}

%% file: tables.tex
\usepackage{caption}
\usepackage{subcaption}
\usepackage{pifont} 
\usepackage{multirow, overpic, textpos, array}
\usepackage{makecell}
\usepackage{footnote}

\newlength\savewidth\newcommand\shline{\noalign{\global\savewidth\arrayrulewidth
  \global\arrayrulewidth 1pt}\hline\noalign{\global\arrayrulewidth\savewidth}}
\newcommand{\tablestyle}[2]{\setlength{\tabcolsep}{#1}\renewcommand{\arraystretch}{#2}\centering\footnotesize}
\renewcommand{\paragraph}[1]{\vspace{1.25mm}\noindent\textbf{#1}}
\newcommand\blfootnote[1]{\begingroup\renewcommand\thefootnote{}\footnote{#1}\addtocounter{footnote}{-1}\endgroup}

\newcolumntype{x}[1]{>{\centering\arraybackslash}p{#1pt}}
\newcolumntype{y}[1]{>{\raggedright\arraybackslash}p{#1pt}}
\newcolumntype{z}[1]{>{\raggedleft\arraybackslash}p{#1pt}}

\newcommand{\app}{\raise.17ex\hbox{$\scriptstyle\sim$}}

\newcommand{\x}{{\times}}

\newcommand{\headrule}{\toprule}

\usepackage{color, colortbl}
\definecolor{Gray}{gray}{0.9}
\definecolor{baselinecolor}{gray}{0.9}

\newcommand{\TableComparisonToOtherSparseDetectors}{
    \begin{table*}[t]
        \footnotesize
        \setlength{\tabcolsep}{3pt}
        \renewcommand\arraystretch{1}
        \centering
        \resizebox{\textwidth}{!}{
        \begin{tabular}{l|l|l|l}
                        & adaptive to decode locations?                                        & adaptive to decode content?                  & extra networks before the query decoder\textsuperscript{1}?                             \\
        \shline
        DETR~\cite{DBLP:conf/eccv/CarionMSUKZ20}            & {\em yes}, multi-head attention aggregation                        & {\em no}, linear projection                    & TransformerEncoder            \\
        Deformable DETR~\cite{DBLP:journals/corr/abs-2010-04159} & {\em yes}, multi-scale multi-head adaptive sampling & {\em no}, linear projection\textsuperscript{2}                   & Multi-scale DeformTransEncoder   \\
        Sparse R-CNN~\cite{DBLP:journals/corr/abs-2011-12450}    & {\em restricted}, RoIAlign~\cite{DBLP:conf/iccv/HeGDG17}                                                    & {\em partially yes}, adaptive point-wise conv.     &FPN
                                  \\
        AdaMixer (ours) & {\em yes}, adaptive 3D sampling                            & {\em yes}, adaptive channel and spatial mixing & linear projection to form 3D feature space                           
        \end{tabular}
        }

        \vspace{-0.4em}
        \caption{\textbf{Comparisons of the adaptability of decoders} across different query-based object detectors. \textsuperscript{1}We specify trainable networks introduced after pre-trained backbones before the query decoder.~\textsuperscript{2}We regard the softmax aggregation in deformable attention as one step in decoding locations as the softmax weights normalize to one.}
        \label{tab:comparsionAdaptability}
        \vspace{-0.5em}
        \end{table*}
}

\newcommand{\TableSOTA}{
\begin{table*}[t]
    \centering
    \small
    \renewcommand\arraystretch{1.0}
    \setlength{\tabcolsep}{4pt}
    \begin{tabular}{l|c|c|c|c|llllll}
        detector & backbone & {encoder/pyramid net} & { \#epochs} & { GFLOPs} & AP  & AP$_{50}$ & AP$_{75}$ & AP$_{s}$ & AP$_m$ & AP$_l$  \\
        \shline
        DETR~\cite{DBLP:conf/eccv/CarionMSUKZ20}
        & ResNet-50-DC5 & TransformerEnc & 500 & 187 & 43.3 & 63.1 & 45.9 & 22.5 & 47.3 & 61.1  \\
        SMCA~\cite{DBLP:journals/corr/abs-2101-07448}
        & ResNet-50 & TransformerEnc & 50 & 152 & 43.7 & 63.6 & 47.2 & 24.2 & 47.0 & 60.4 \\
        Deformable DETR~\cite{DBLP:journals/corr/abs-2010-04159}
        & ResNet-50 & DeformTransEnc & 50 & 173 & 43.8 & 62.6 & 47.7 & 26.4 & 47.1 & 58.0  \\
        Sparse R-CNN~\cite{DBLP:journals/corr/abs-2011-12450}
        & ResNet-50 & FPN & \textbf{36} & 174  & 45.0 & 63.4 & 48.2 & 26.9 & 47.2 & 59.5  \\
        Efficient DETR~\cite{DBLP:journals/corr/abs-2104-01318}
        & ResNet-50 & DeformTransEnc & \textbf{36} & 210 & 45.1 & 63.1 & 49.1 & 28.3 & 48.4 & 59.0 \\
        Conditional DETR~\cite{DBLP:journals/corr/abs-2108-06152}
        & ResNet-50-DC5 & TransformerEnc & 108 & 195 & 45.1 & 65.4 & 48.5 & 25.3 & 49.0 & \textbf{62.2}\\
        Anchor DETR~\cite{DBLP:journals/corr/abs-2109-07107}
        & ResNet-50-DC5 & DecoupTransEnc & 50 & 151 & 44.2 & 64.7 & 47.5 & 24.7 & 48.2 & 60.6 \\

        \textbf{AdaMixer (ours)}
        & ResNet-50 & - & \textbf{12} & 132 & 44.1 & 63.1 & 47.8 & 29.5 & 47.0 & 58.8  \\
        \textbf{AdaMixer (ours)}
        & ResNet-50 & - & \textbf{24} & 132 & 46.7 & 65.9 & 50.5 & 29.7 & 49.7 & 61.5   \\
        \textbf{AdaMixer (ours)}
        & ResNet-50 & - & \textbf{36} & 132 & \textbf{47.0} & \textbf{66.0} & \textbf{51.1} & \textbf{30.1} & \textbf{50.2} & {61.8}  \\

        \hline
        DETR~\cite{DBLP:conf/eccv/CarionMSUKZ20}
        & ResNet-101-DC5 & TransformerEnc & 500 & 253 & 44.9 & 64.7 & 47.7 & 23.7 & 49.5 & 62.3 \\
        SMCA~\cite{DBLP:journals/corr/abs-2101-07448}
        & ResNet-101 & TransformerEnc & 50 & 218 & 44.4 & 65.2 & 48.0 & 24.3 & 48.5 & 61.0 \\
        Sparse R-CNN~\cite{DBLP:journals/corr/abs-2011-12450}
        & ResNet-101 & FPN & \textbf{36} & 250 & 46.4 & 64.6 & 49.5 & 28.3 & 48.3 & 61.6 \\
        Efficient DETR~\cite{DBLP:journals/corr/abs-2104-01318}
        & ResNet-101 & DeformTransEnc & \textbf{36} & 289 & 45.7 & 64.1 & 49.5 & 28.2 & 49.1 & 60.2 \\
        Conditional DETR~\cite{DBLP:journals/corr/abs-2108-06152}
        & ResNet-101-DC5 & TransformerEnc & 108 & 262 & 45.9 & 66.8 & 49.5 & 27.2 & 50.3 & 63.3 \\

        \textbf{AdaMixer (ours)}
        & ResNet-101 & - & \textbf{36} & 208 & \textbf{48.0} & \textbf{67.0} & \textbf{52.4} & \textbf{30.0} & \textbf{51.2} & \textbf{63.7} \\
        \hline
        \textbf{AdaMixer (ours)}
        & ResNeXt-101-DCN & - & \textbf{36} & 214 & \textbf{49.5} & \textbf{68.9} & \textbf{53.9} & \textbf{31.3} & \textbf{52.3} & \textbf{66.3} \\
        \textbf{AdaMixer (ours)}
        & Swin-S & - & \textbf{36} & 234 & \textbf{51.3} & \textbf{71.2} & \textbf{55.7} & \textbf{34.2} & \textbf{54.6} & \textbf{67.3} \\
        \end{tabular}
    \vspace{-1.0em}
    \caption{\textbf{Different query-based detector performance} on COCO \texttt{minival} set with \textbf{longer training scheme} and single scale testing.
    }
    \label{tab:longscheme}
    \vspace{-1.4em}
\end{table*}
}

\newcommand{\TableOneTimesSchedule}{
\begin{table}[t]
    \centering
    \renewcommand\arraystretch{1.0}
    \resizebox{0.49\textwidth}{!}{
    \begin{tabular}{l|c|llllll}
        detector & \footnotesize epochs & AP  & AP$_{50}$ & AP$_{75}$ & AP$_{s}$ & AP$_m$ & AP$_l$ \\
        \shline
        \footnotesize FCOS~\cite{DBLP:conf/iccv/TianSCH19}
            & 12 & 38.7 & 57.4 & 41.8 & 22.9 & 42.5 & 50.1 \\
        \footnotesize Cascade R-CNN~\cite{DBLP:conf/cvpr/CaiV18}
            & 12 & 40.4 & 58.9 & 44.1 & 22.8 & 43.7 & 54.0 \\
        \footnotesize GFocalV2~\cite{DBLP:conf/cvpr/LiW0LT021}
            & 12 & 41.1 & 58.8 & 44.9 & 23.5 & 44.9 & 53.3 \\
        \footnotesize BorderDet~\cite{DBLP:conf/eccv/QiuMLLS20}
            & 12 & 41.4 & 59.4 & 44.5 & 23.6 & 45.1 & 54.6 \\
        \footnotesize Dynamic Head~\cite{DBLP:conf/cvpr/DaiCX0LY021}
            & 12 & 42.6 & 60.1 & \textbf{46.4} & \textbf{26.1} & \textbf{46.8} & 56.0 \\
        \footnotesize DETR~\cite{DBLP:conf/eccv/CarionMSUKZ20}
            & 12 & 20.0 & 36.2 & 19.3 & 6.0  & 20.5 & 32.2 \\
        \footnotesize Deformable DETR~\cite{DBLP:journals/corr/abs-2010-04159}
            & 12 & 35.1 & 53.6 & 37.7 & 18.2 & 38.5 & 48.7 \\
        \footnotesize Sparse R-CNN~\cite{DBLP:journals/corr/abs-2011-12450}
            & 12 & 37.9 & 56.0 & 40.5 & 20.7 & 40.0 & 53.5 \\
        \textbf{AdaMixer} ($N$=100)
            & 12 & \textbf{42.7} & \textbf{61.5} & {45.9} & {24.7} & {45.4} & \textbf{59.2} \\
        \hline
        \textbf{AdaMixer} ($N$=300)
            & 12 & \textbf{44.1} & \textbf{63.4} & \textbf{47.4} & \textbf{27.0} & \textbf{46.9} & \textbf{59.5} \\
        \textbf{AdaMixer} ($N$=500)
            & 12 & \textbf{45.0} & \textbf{64.2} & \textbf{48.6} & \textbf{27.9} & \textbf{47.8} & \textbf{61.1} \\
        \end{tabular}
        }
    \vspace{-0.4em}
    \caption{\textbf{$\mathbf{1}\times$ training scheme} performance on COCO \texttt{minival} set with different detectors and ResNet-50 as backbone.
    }
    \label{table:onetimeschedule}
    \vspace{-1.0em}
\end{table}
}

\newcommand{\TableConfiguration}{

\vspace{-0.3em}
\begin{table}[h!]
    \centering
    \footnotesize
    \setlength{\tabcolsep}{3pt}
    \renewcommand\arraystretch{1.2}
    \resizebox{0.48\textwidth}{!}{
    \begin{tabular}{ccccccccc}
        \makecell[c]{query dim\\ $d_{q}$} & \makecell[c]{feat. maps\\ used} & \makecell[c]{feat. dim\\ $d_{\rm feat}$} & \makecell[c]{\#stages in\\decoder} & \makecell[c]{\#groups\\ $g$} & $C$ & {$P_{\rm in}$} & {$P_{\rm out}$} \\
        \shline
        256 & $C_{2}\sim C_{5}$ & 256 & 6 & 4 & 64 & 32 & 128 \\
    \end{tabular}
    }
    \vspace{-1.2em}
    \captionof{table}{\textbf{Default configuration} of our AdaMixer detector.
    }\label{tab:configuration}
\end{table}

\vspace{-0.5em}
}

\newcommand{\TableGroup}{
\newcolumntype{a}{>{\columncolor{Gray}}c}
\begin{table}[h!]
    \centering
    \footnotesize
    \setlength{\tabcolsep}{10pt}
    \renewcommand\arraystretch{1.0}
    \begin{tabular}{c|ccac}
        $g$ & 1 & 2 & 4 & 8 \\
        \shline
        AP & 42.5 & \textbf{42.8} & 42.7 & 41.9  \\
        FLOPs  & 111G & 106G & \textbf{104G} & 106G \\
        Params & 191M & 148M & \textbf{135M} & 149M
    \end{tabular}
    \vspace{-1.2em}
    \captionof{table}{\textbf{Grouping} sampling and mixing with $g$ groups.
    }\label{tab:group}
\end{table}
}

\newcommand{\TableAblationOnAdaptiveDesign}{
        \multicolumn{2}{c}{adaptive} & \multirow{2}{*}{AP} & \multirow{2}{*}{AP$_{50}$} & \multirow{2}{*}{AP$_{75}$} & \multirow{2}{*}{AP$_{s}$} & \multirow{2}{*}{AP$_{m}$} & \multirow{2}{*}{AP$_{l}$} \\
        loc. & cont. & \\
        \shline
        & &
            35.7 & 55.2 & 37.8 & 20.1 & 38.1 & 48.8 \\
        \checkmark & &
            37.3 & 55.8 & 39.7 & 20.7 & 40.1 & 50.9 \\
        & \checkmark &
            40.4 & 60.5 & 43.4 & 23.0 & 42.5 & 56.7 \\
        \rowcolor{Gray}
        \checkmark & \checkmark &
            \textbf{42.7} & \textbf{61.5} & \textbf{45.9} & \textbf{24.7} & \textbf{45.4} & \textbf{59.2} \\
}

\newcommand{\TableAblationOnPin}{
        $P_{\rm in}$ & AP  & AP$_{50}$ & AP$_{75}$ & AP$_{s}$ & AP$_m$ & AP$_l$ \\
        \headrule
        8 & 41.2 & 60.3 & 44.1 & 24.0 & 43.9 & 57.2 \\
        16 & 41.8 & 60.9 & 44.5 & 24.5 & 44.6 & 58.4 \\
        \rowcolor{Gray}
        32 & \textbf{42.7} & \textbf{61.5} & 45.9 & 24.7 & 45.4 & 59.2 \\
        64 & \textbf{42.7} & \textbf{61.5} & \textbf{46.1} & \textbf{24.9} & \textbf{45.5} & \textbf{59.3} \\
}

\newcommand{\TableAblationOnPout}{
        $P_{\rm out}$ & AP  & AP$_{50}$ & AP$_{75}$ & AP$_{s}$ & AP$_m$ & AP$_l$ \\
        \headrule
        32 & 41.1 & 60.0 & 44.0 & 24.5 & 43.6 & 57.2 \\
        64 & 42.1 & 61.2 & 45.0 & 24.0 & 44.8 & 57.8 \\
        \rowcolor{Gray}
        128 & \textbf{42.7} & \textbf{61.5} & \textbf{45.9} & \textbf{24.7} & \textbf{45.4} & \textbf{59.2} \\
        256 & 42.4 & 61.4 & 45.5 & 24.4 & 45.0 & 58.7  \\
}

\newcommand{\TableAblationOnExtraNecks}{
        pyramid & AP  & AP$_{50}$ & AP$_{75}$ & AP$_{s}$ & AP$_m$ & AP$_l$ \\
        \shline
        FPN~\cite{DBLP:conf/cvpr/LinDGHHB17} & 42.1 & 61.0 & 45.0 & 24.1 & 44.8 & 58.7 \\
        PAFPN~\cite{DBLP:conf/cvpr/LiuQQSJ18} & 41.7 & 60.5 & 44.7 & 23.5 & 44.6 & 58.7 \\
        \rowcolor{Gray}
        - & \textbf{42.7} & \textbf{61.5} & \textbf{45.9} & \textbf{24.7} & \textbf{45.4} & \textbf{59.2} \\
}

\newcommand{\TableAblationMixDesign}{
        \multicolumn{2}{c}{mixing} & AP  & AP$_{50}$ & AP$_{75}$ & AP$_{s}$ & AP$_m$ & AP$_l$ \\
        \shline
        ACM & ACM & 41.5 & 60.5 & 44.3 & 23.5 & 44.1 & 57.4 \\
        ASM & ASM & 39.8 & 58.8 & 42.6 & 22.8 & 42.4 & 56.1 \\
        \rowcolor{Gray}
        ACM & ASM & \textbf{42.7} & \textbf{61.5} & \textbf{45.9} & \textbf{24.7} & \textbf{45.4} & \textbf{59.2} \\
        ASM & ACM & 41.5 & 60.4 & 44.5 & 23.9 & 44.4 & 57.1 \\
}

\newcommand{\TableAblationOnPositionalMHSA}{
        \multicolumn{2}{c}{pos. inf.} & \multirow{2}{*}{AP} & \multirow{2}{*}{AP$_{50}$} & \multirow{2}{*}{AP$_{75}$} & \multirow{2}{*}{AP$_{s}$} & \multirow{2}{*}{AP$_{m}$} & \multirow{2}{*}{AP$_{l}$} \\
        sinus. & IoF & \\
        \headrule
        & &
            41.2 & 59.6 & 44.2 & 23.6 & 43.5 & 57.9 \\
        \checkmark & &
            41.5 & 59.9 & 44.3 & 23.6 & 44.0 & 57.8 \\
        & \checkmark &
            42.2 & 61.2 & 45.0 & \textbf{24.8} & 45.1 & 58.8 \\
        \rowcolor{Gray}
        \checkmark & \checkmark &
            \textbf{42.7} & \textbf{61.5} & \textbf{45.9} & {24.7} & \textbf{45.4} & \textbf{59.2} \\
}

\newcommand{\TableAblationStudiesIntegrated}{
    \begin{table*}[t]
        \vspace{-.2em}
        \centering
        \subfloat[\small
        \textbf{Adaptability of decoding} sampling locations and sampled content.
        \label{tab:adaptabilityDecoding}
        ]{
        \centering
        \begin{minipage}{0.3\linewidth}{\begin{center}
        \tablestyle{1pt}{1.1}
        \small
        \begin{tabular}{x{14}x{18}x{18}x{18}x{18}x{18}x{18}x{18}}
            \TableAblationOnAdaptiveDesign
        \end{tabular}
        \end{center}}\end{minipage}
        }
        \hspace{1em}
        \subfloat[\small
        \textbf{Design} in our adaptive mixing procedure.
        \label{tab:mixingDesign}
        ]{
        \begin{minipage}{0.3\linewidth}{\begin{center}
        \tablestyle{0.6pt}{1.1}
        \small
        \begin{tabular}{x{18}x{22}x{18}x{18}x{18}x{18}x{18}x{18}}
            \TableAblationMixDesign
        \end{tabular}
        \end{center}}\end{minipage}
        }
        \hspace{1em}
        \subfloat[\small
        \textbf{Extra pyramid networks} after the backbone?
        \label{tab:extraNeck}
        ]{
        \begin{minipage}{0.3\linewidth}{\begin{center}
        \tablestyle{1pt}{1.1}
        \small
        \begin{tabular}{x{42}x{18}x{18}x{18}x{18}x{18}x{18}x{18}}
            \TableAblationOnExtraNecks
        \end{tabular}\vspace{1.35em}
        \end{center}}\end{minipage}
        }\\\vspace{-0.5em}
        \subfloat[\small
        \textbf{Sampling points $P_{\rm in}$} per group.
        \label{tab:pin}
        ]{
        \centering
        \begin{minipage}{0.3\linewidth}{\begin{center}
        \tablestyle{1pt}{1.1}
        \small
        \begin{tabular}{x{24}x{18}x{18}x{18}x{18}x{18}x{18}x{18}}
            \TableAblationOnPin
        \end{tabular}
        \end{center}}\end{minipage}
        }
        \hspace{1em}
        \subfloat[\small
        \textbf{Spatial mixing out patterns $P_{\rm out}$} per group.
        \label{tab:pout}
        ]{
        \centering
        \begin{minipage}{0.3\linewidth}{\begin{center}
            \tablestyle{1pt}{1.1}
        \small
        \begin{tabular}{x{24}x{18}x{18}x{18}x{18}x{18}x{18}x{18}}
            \TableAblationOnPout
        \end{tabular}
        \end{center}}\end{minipage}
        }
        \hspace{1em}
        \subfloat[\small
        \textbf{Position information} in self-attention between queries.
        \label{tab:positionalInformation}
        ]{
        \centering
        \begin{minipage}{0.3\linewidth}{\begin{center}
        \tablestyle{1pt}{1.1}
        \small
        \begin{tabular}{x{18}x{18}x{18}x{18}x{18}x{18}x{18}x{18}}
            \TableAblationOnPositionalMHSA
        \end{tabular}
        \end{center}}\end{minipage}
        }
        \\
        \vspace{-0.4em}
        \caption{\textbf{AdaMixer ablation experiments} with ResNet-50 on MS COCO \texttt{minival} set. Default choice for our model is colored \colorbox{Gray}{gray}
        }
        \label{tab:ablations} \vspace{-1.0em}
        \end{table*}
}

\newcommand{\TableTestDev}{
\begin{table}[h]
    \centering
    \small
    \renewcommand\arraystretch{1.0}
    \setlength{\tabcolsep}{3pt}
    \begin{tabular}{c|llllll}
        backbone & AP  & AP$_{50}$ & AP$_{75}$ & AP$_{s}$ & AP$_m$ & AP$_l$  \\
        \shline
        ResNet-50 & 47.2 & 66.3 & 51.4 & 28.3 & 49.3 & 60.6  \\

        ResNet-101 &48.1 & 67.3 & 52.4 & 28.3 & 50.6 & 62.1 \\

        ResNeXt-101-DCN  & 49.3 & 68.8 & 53.7 & 30.0 & 51.6 & 64.0 \\

        Swin-S & 51.3 & 71.3 & 55.9 & 31.4 & 53.9 & 66.3
        \end{tabular}
    \caption{AdaMixer performance on COCO \texttt{test-dev} set with \textbf{longer training scheme} (36 epochs training and 300 queries) and single model single scale testing.
    }
    \label{tab:testdev}
\end{table}
}

\newcommand{\TableAblationOnSamplingContent}{
    sampling & AP  & AP$_{50}$ & AP$_{75}$ & AP$_{s}$ & AP$_m$ & AP$_l$ \\
        \shline
        only C$_3$ feature
            & 26.2 & 42.0 & 27.3 & 15.8 & 28.7 & 34.1 \\
        only C$_4$ feature
            & 38.3 & 57.2 & 41.0 & 20.0 & 41.9 & 54.1\\
        only C$_5$ feature
            & 37.8 & 58.3 & 39.5 & 18.0 & 41.2 & 51.7 \\
        RoIAlign
            & 37.2 & 58.5 & 39.0 & 19.0 & 39.3 & 55.6 \\
        \hline
        A2DS
            & 41.3 & 61.0 & 44.4 & 23.3 & 43.8 & 57.8 \\
        \rowcolor{Gray} 
        A3DS & \textbf{42.7} & \textbf{61.5} & \textbf{45.9} & \textbf{24.7} & \textbf{45.4} & \textbf{59.2} \\
}

\newcommand{\TableAblationThreeDSampling}{
    \begin{table}[h]
        \vspace{-.2em}
        \centering
        {\begin{center}
        \tablestyle{1pt}{1.0}
        \small
        \begin{tabular}{x{64}x{24}x{24}x{24}x{24}x{24}x{24}}
            \TableAblationOnSamplingContent
        \end{tabular}
        \end{center}}
        \caption{\textbf{Different feature sampling} performance on MS COCO \texttt{minival} set with $1\times$ training scheme. The ``A3DS'' is abbreviation for \textbf{A}daptive \textbf{3D} \textbf{S}ampling used in AdaMixer. The ``A2DS'' is the 2D variant of adaptive 3D sampling, where it disables learning adaptive z-axis offsets, namely, $\Delta z_{(\cdot)}=0$ through training and inference. Note that A2DS is still with multiple feature maps and $z$ of a query still decides how feature maps are aggregated. The RoIAlign is performed to extract features from feature maps individually according to the bounding box indicated by a query. The row ``only C$_{(\cdot)}$ feature'' variants use a single feature map to perfom adaptive 2D sampling. No FPNs used throughout this table.}
        \label{tab:sampling}
    \end{table}
}

\newcommand{\TableModel}{
\begin{table}[h]
    \centering
    \footnotesize
    \renewcommand\arraystretch{1.0}
    \setlength{\tabcolsep}{1pt}
    \begin{tabular}{lc|cc|cccc}
        detector & backbone & AP$_\text{val}$  & AP$_\text{test}$ & \makecell[c]{\#Param\\(M)} & \makecell[c]{flops\\(G)} & FPS & \makecell[c]{training\\ hours}  \\
        \shline
        Faster R-CNN & R-50
        & 41.8 & - & 42 & 207 & 17 & $\sim$16 \\
        DETR & R-50-DC5
        & 43.3 & - & 41 & 187 & 11 & $\sim$204  \\
        Deformable DETR++ & R-50
        & 46.2 & 46.9 & 40 & 173 & 12 & $\sim$106  \\
        Sparse R-CNN & R-50
        & 45.0 & - & 110 & 174 & 16 & $\sim$30 \\
        
        {AdaMixer} & R-50 
        & 47.0 & 47.2 & 139 & 132 & 15 & 29 \\
        \hline

        {AdaMixer} & R-101
        & 48.0 & 48.1 & 158 & 208 & 12 & 37 \\

        {AdaMixer} & X-101-DCN
        & 49.5 & 49.3 & 160 & 214 & 8  & 65 \\

        {AdaMixer} & Swin-S
        & 51.3 & 51.3 & 164 & 234 & 10 & 51
        \end{tabular}
    \caption{
        Training and inference details about different models. AdaMixer models here are ones in Table~\ref{tab:longscheme} of the paper with 300 queries. AP$_\text{val}$ and AP$_\text{test}$ are the average precision on COCO \texttt{minival} and \texttt{test-dev} set, respectively. FPS is calculated on a single Nvidia V100 card. Training hours for other detectors are estimated on 8 V100 cards when obtaining the speed of a few training iterations.
    }
    \label{tab:model}
\end{table}
}

%% file: figures.tex
\usepackage{caption}
\usepackage{subcaption}

\newcommand{\FigureCurve}{
    \begin{figure}[t]
        \centering
        \includegraphics[width=0.5\textwidth]{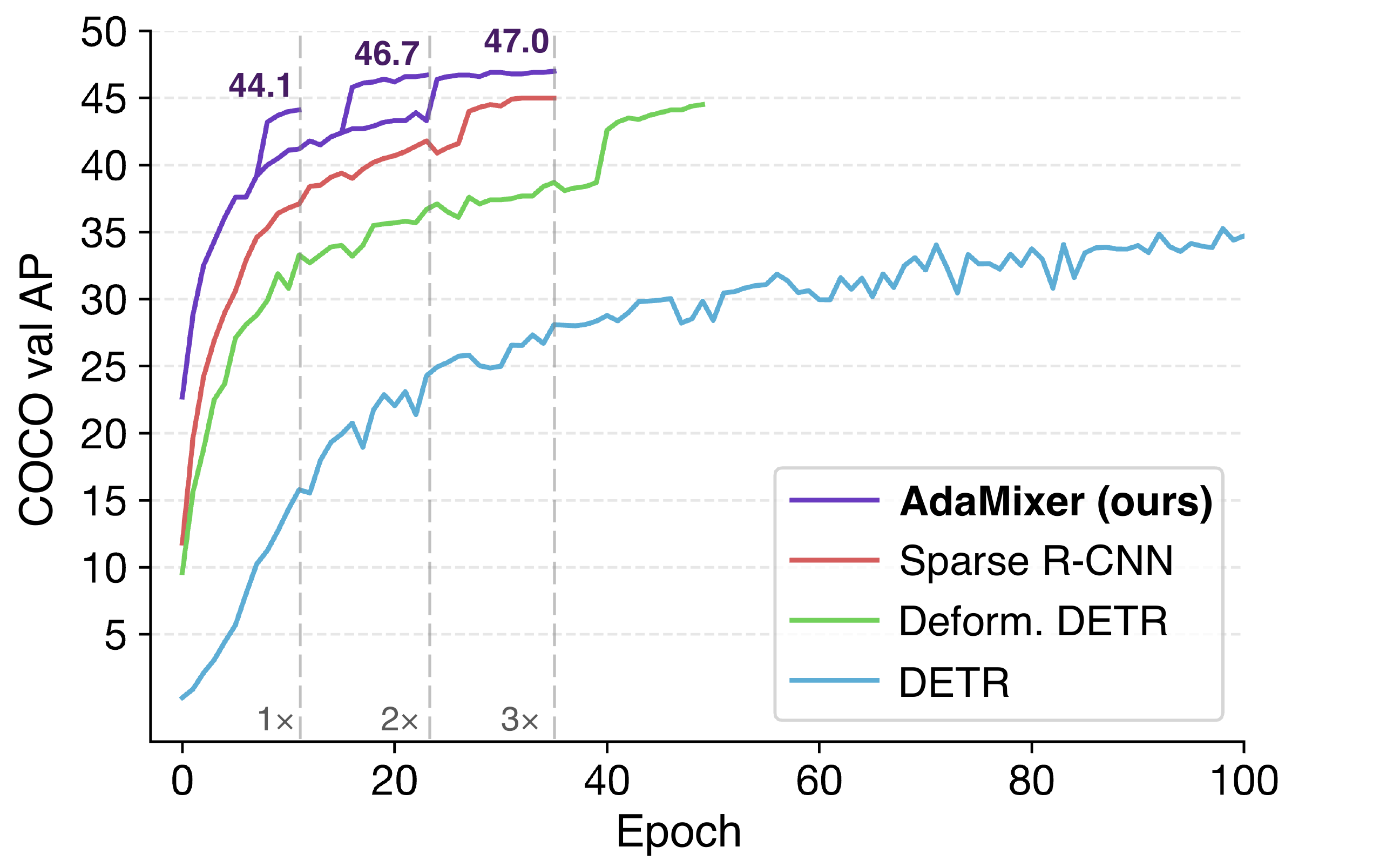}
        \caption{\textbf{Convergence curves} of our AdaMixer, DETR, Deformable DETR and Sparse R-CNN with ResNet-50 as the backbone on MS COCO minival set.}
        \label{fig:curve}
        \vspace{-1.5em}
    \end{figure}
        
}

\newcommand{\FigureStructure}{
    \begin{figure*}[t]
        \centering
        \includegraphics[width=0.9\textwidth]{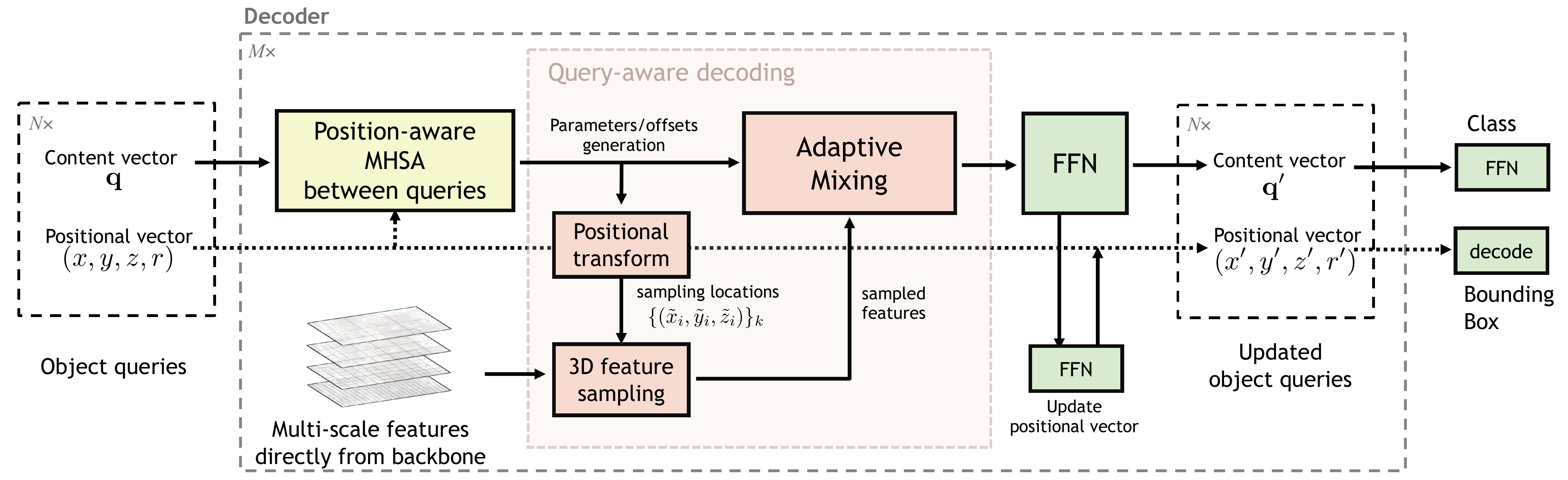}
        \caption{\textbf{Our decoder structure} of the AdaMixer. There are two operator streams on a query: one on its content vector $\mathbf{q}$ (the solid horizontal line) and one on its positional vector $(x,y,z,r)$ (the dashed horizontal line). Each operator on the content vector in the decoder is followed by a  residual addition and LayerNorm.
        }
        \label{fig:structure}
        \vspace{-0.5em}
    \end{figure*}
        
}

\newcommand{\FigureSamplerAndMixing}{
    \begin{figure*}
        \centering
        \begin{minipage}[t]{.33\textwidth}
          \centering
          \includegraphics[width=.999\linewidth]{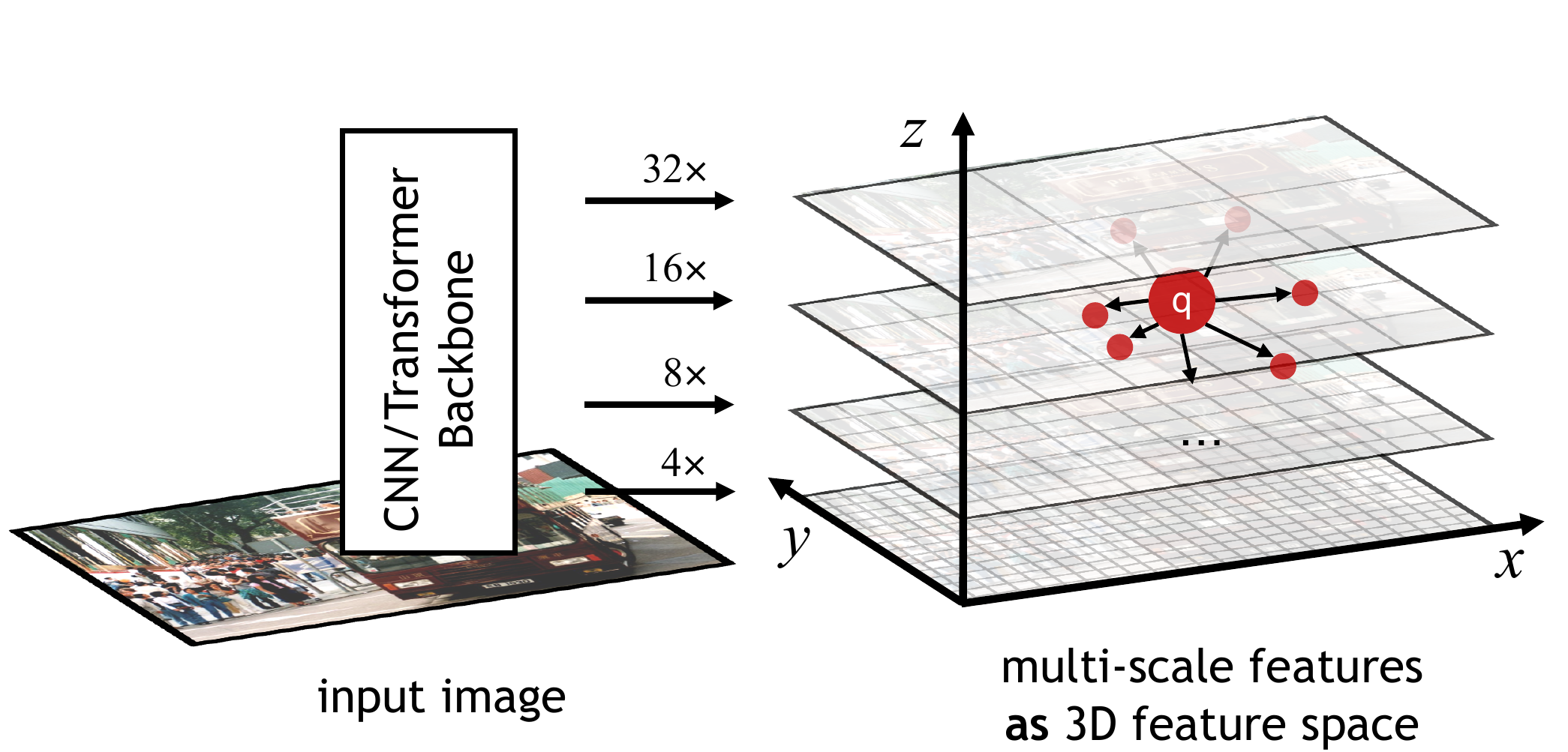}
          \captionof{figure}{\textbf{3D feature sampling process.} A query first obtains sampling points in the 3D feature space and then perform 3D interpolation on these sampling points.}
          \label{fig:sampler}
        \end{minipage}\hspace{0.5em}
        \begin{minipage}[t]{.63\textwidth}
          \centering
          \includegraphics[width=.999\linewidth]{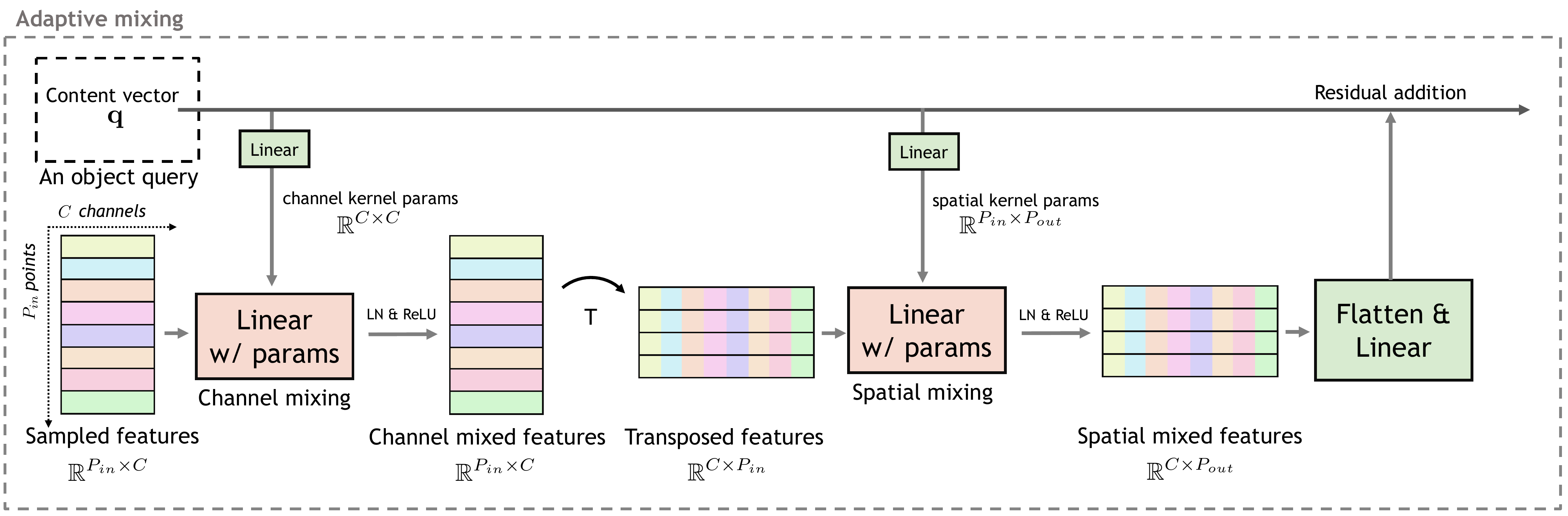}
          \captionof{figure}{
            \textbf{Adaptive mixing procedure} between an object query and sampled features. The object query first generates adaptive mixing weights and then apply these weights to mix sampled features in the channel and spatial dimension. Note that for clarity, we demonstrate adaptive mixing for one sampling group.
          }
          \label{fig:mixing}
        \end{minipage}
        \end{figure*}
}

\newcommand{\FigSuppA}{
    \begin{figure*}[b]
        \centering
        \includegraphics[width=0.85\textwidth]{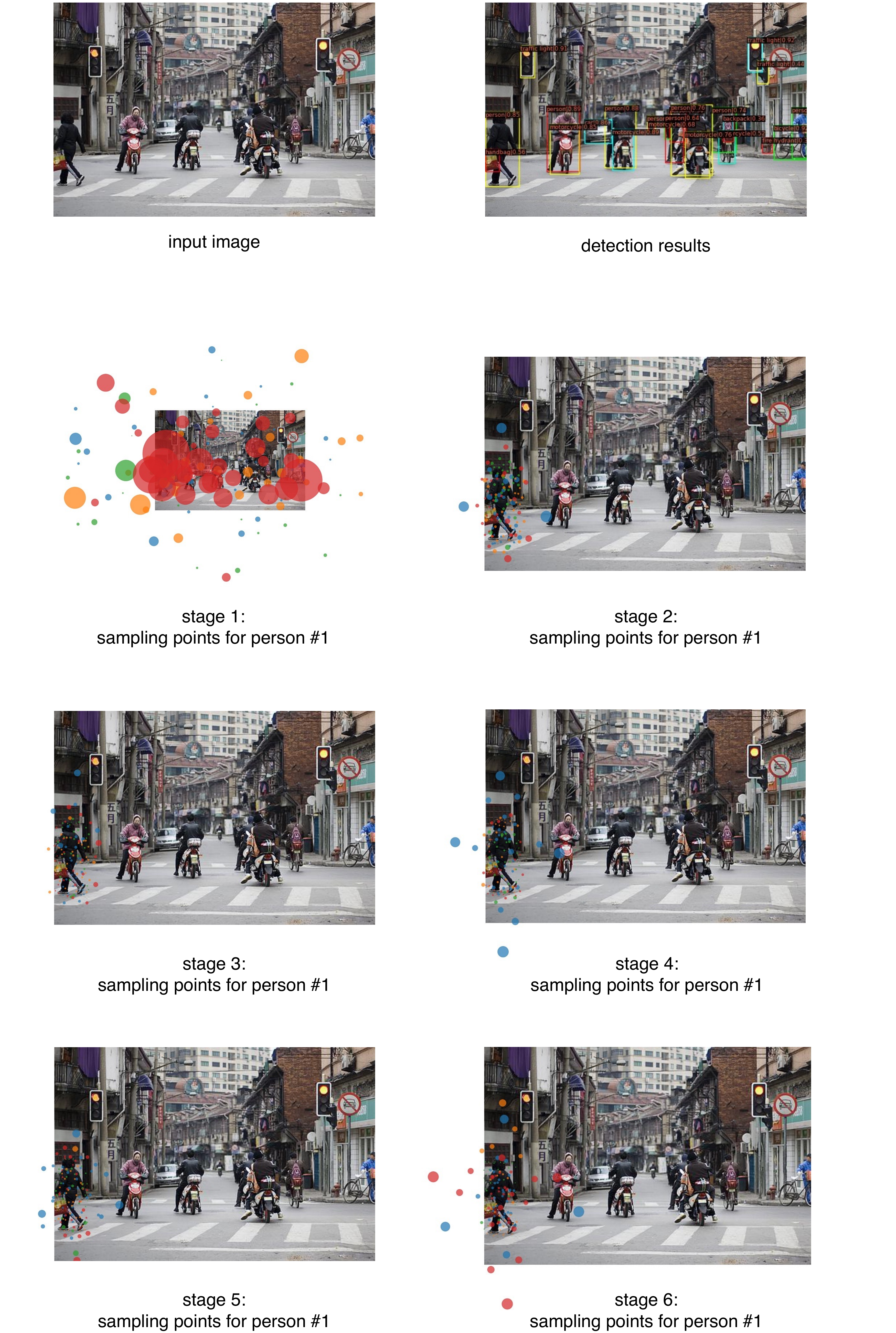}
        \caption{Visualizations of detection results and sampling points of AdaMixer with ResNet-50 as the backbone.}
        \label{fig:a}
        \vspace{-1.5em}
    \end{figure*} 
}

\newcommand{\FigSuppB}{
    \begin{figure*}[b]
        \centering
        \includegraphics[width=0.85\textwidth]{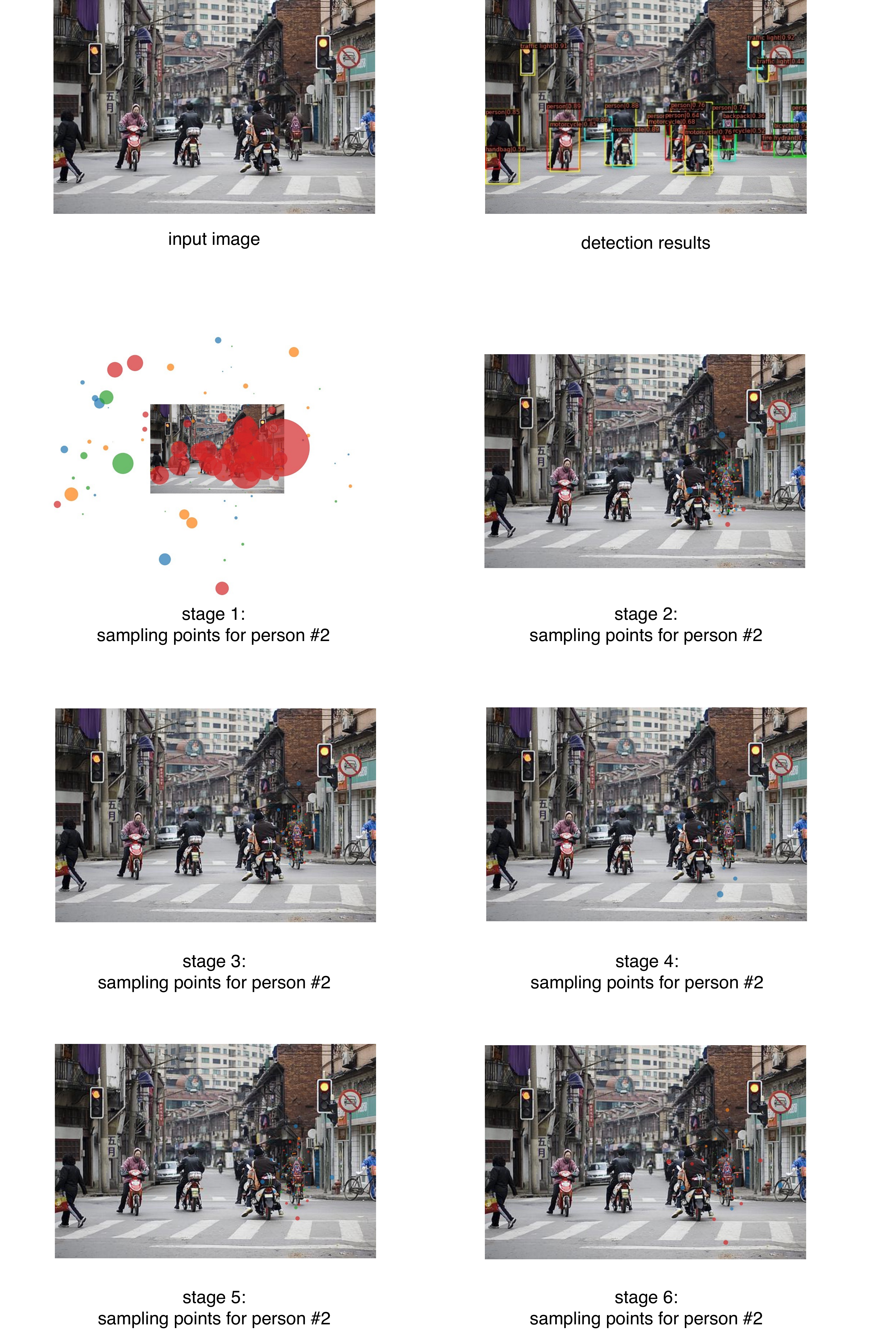}
        \caption{Visualizations of detection results and sampling points of AdaMixer with ResNet-50 as the backbone (cont'd).}
        \label{fig:b}
        \vspace{-1.5em}
    \end{figure*} 
}

%% file: related.tex
\section{Related Work}
\noindent\textbf{Dense object detectors.}
The dense paradigm of object detectors dates back to sliding window-based approaches~\cite{DBLP:conf/cvpr/ViolaJ01, DBLP:conf/cvpr/FelzenszwalbGM10, DBLP:journals/pami/FelzenszwalbGMR10}, which involve exhaustive classification over space and scales due to the assumption of potential objects emerging uniformly and densely w.r.t. spatial locations in an image.
This assumption about natural images remains effective in the deep learning era for its power to cover potential objects~\cite{DBLP:journals/corr/SermanetEZMFL13}. Prevalent object detectors in the past few years, \eg, one-stage detectors~\cite{DBLP:conf/mm/YuJWCH16, DBLP:conf/eccv/LiuAESRFB16, DBLP:conf/cvpr/RedmonDGF16, DBLP:journals/corr/abs-1804-02767, DBLP:conf/iccv/TianSCH19, DBLP:journals/corr/abs-1904-07850}, multiple-stage detectors~\cite{DBLP:conf/nips/RenHGS15, DBLP:conf/iccv/HeGDG17, DBLP:conf/cvpr/CaiV18, DBLP:conf/cvpr/ChenPWXLSF0SOLL19, DBLP:conf/cvpr/PangCSFOL19} or point-based methods~\cite{DBLP:journals/corr/abs-1904-07850, DBLP:conf/eccv/LawD18, DBLP:conf/cvpr/ZhouZK19, DBLP:conf/iccv/DuanBXQH019}, are also rooted in this dense assumption in either region proposal networks or entire object detector architectures.
They apply dense priors, such as anchors or anchor points, upon the feature map to exhaustively find foreground objects or directly classify them.

\noindent\textbf{Query-based object detectors.}
Recently transformer-based detector, DETR~\cite{DBLP:conf/eccv/CarionMSUKZ20}, formulates object detection as a direct set prediction task and achieves promising performance. DETR predicts a set of objects by attending queries to the feature map with the transformer decoder.
The original architecture of DETR is simply based on the Transformer~\cite{DBLP:conf/nips/VaswaniSPUJGKP17}, which contains the multi-layer attentional encoder and decoder.
The training for set prediction in DETR is based on the bipartite matching between the predictions and ground-truth objects.
While DETR outperforms competitive Faster R-CNN baselines, it still suffers from limited spatial resolution, poor small object detection performance, and slow training convergence.
There have been several work to tackle these issues. Deformable DETR~\cite{DBLP:journals/corr/abs-2010-04159} considers the shift-equivalence in natural images and introduces the multi-scale deformable family of attention operators in both encoders and decoders of DETR. SMCA~\cite{DBLP:journals/corr/abs-2101-07448}, Conditional DETR~\cite{DBLP:journals/corr/abs-2108-06152}, and Anchor DETR~\cite{DBLP:journals/corr/abs-2109-07107} explicitly model the positional attention for foreground objects to fasten the convergence. Efficient DETR~\cite{DBLP:journals/corr/abs-2104-01318} bridges the dense prior to queries in DETR to improve the performance.
Sparse R-CNN~\cite{DBLP:journals/corr/abs-2011-12450} brings the query-based paradigm of DETR to Cascade R-CNN~\cite{DBLP:conf/cvpr/CaiV18} and introduces the dynamic instance interaction head and its query-adaptive point-wise convolution, to effectively cast queries to potential objects.

Our AdaMixer generally follows this research line of using queries to attend features for object detection. However, we improve the query-based object detection paradigm from a new perspective: the adaptability of decoding queries across images. Specifically, we focus on how to make decoding scheme on queries more adaptive to the content of images from both semantic and spatial aspects. We present adaptive 3D feature sampling and adaptive content decoding to improve its flexibility to relate queries with each image. This makes AdaMixer a fast-converging query-based object detector without the introduction of extra feature encoders or explicit pyramid networks.

%% file: method.tex
\section{Approach}
In this paper, we focus on the query decoder in query-based object detectors since the decoder design is essential to casting learned queries to potential objects in each image.
We first revisit decoders in popular query-based object detectors from the perspective of semantic and positional adaptability, and then elaborate on our proposed adaptive query decoder.

\subsection{Object Query Decoder Revisited}\label{sec:revisiting}
\noindent\textbf{Plain attention decoders.}
DETR~\cite{DBLP:conf/eccv/CarionMSUKZ20} applies plain multi-head cross attention between queries and features to cast object queries to potential objects.
As depicted in Table~\ref{tab:comparsionAdaptability}, the cross attention decoder is adaptive to decode sampling locations in the sense that it exploits the relation of object queries and features to aggregate features.
However, the linear transformation of features after aggregation fails to adaptively decode them based on the query.

\noindent\textbf{Deformable multi-scale attention decoders.}
Deformable DETR~\cite{DBLP:journals/corr/abs-2010-04159} improves the ability of decoding sampling locations in plain cross attention in terms of shift equivalence and scale invariance by introducing explicit reference points and multi-scale features. But like DETR, the content decoding of sampled features still remains static by the linear transformation.
Overall, decoders in DETR and Deformable DETR lack the reasoning of aggregated features conditionally on the query and thus limit the semantic adaptability of queries to features. As a result, both of them require stacks of extra attentional encoders to enrich feature semantics.

\noindent\textbf{RoIAlign and dynamic interactive head as decoders.}
Sparse R-CNN~\cite{DBLP:journals/corr/abs-2011-12450}, as the intersection between region-based and query-based detectors, uses the RoIAlign operator and dynamic interactive head as the query decoder.
The dynamic interactive head uses point-wise convolutions, whose kernel is adaptive based on the query, to process RoI features.
This enables the adaptability of queries to RoI features but only partially, in the sense that the adaptive point-wise convolution can not infer adaptive spatial structures from those features to build queries.
Moreover, the sampling locations by RoIAlign operator~\cite{DBLP:conf/iccv/HeGDG17} are restricted inside of the box indicated by a query and a specific level in FPN~\cite{DBLP:conf/cvpr/LinDGHHB17}, which limits positional adaptability and requires explicit pyramid networks for multi-scale modeling.

\noindent\textbf{Summary.}
Given a limited number of queries and varying potential objects across images, an ideal decoder should consider both the semantic and positional adaptability of such queries to the content of images, that is, how to adaptively decode sampling locations and sampled content. This naturally motivates our design of AdaMixer.

\subsection{Our Object Query Definition}
Starting from the object query definition,
we associate two vectors with a query following our semantic and positional view of decoders: one is the content vector $\mathbf{q}$ and the other is the positional vector $(x, y, z, r)$.
This is also in line with \cite{DBLP:journals/corr/abs-2011-12450, DBLP:journals/corr/abs-2109-07107, DBLP:journals/corr/abs-2010-04159} to disentangle the location or the represented bounding box of a query from its content.
The content vector $\mathbf{q}$ is a vector in $\mathbb{R}^{d_{q}}$ and $d_{q}$ is the channel dimension.
The vector $(x, y, z, r)$ describes scaled geometric properties of the bounding box indicated by a query, that is, the x- and y-axis coordinates of its center point and the logarithm of its scale and aspect ratio.
The $x, y, z$ components also directly represent coordinates of a query in the 3D feature space, which will be introduced below.

\noindent\textbf{Decoding the bounding box from a query}. We can simply decode the bounding box from the positional vector. The center $(x_B, y_B)$ and the width and height $w_B$ and $h_B$ of the indicated bounding box can be decoded:
\begin{align}
    &x_{B} = s_{\rm base}\cdot x, \hspace{1.7em}y_{B} = s_{\rm base}\cdot y,\\
    &w_{B} = s_{\rm base}\cdot 2^{z-r}, h_{B} = s_{\rm base}\cdot 2^{z+r},
\end{align}
where $s_{\rm base}$ is the base downsampling stride offset and we set $s_{\rm base}=4$ according to the stride of the largest feature map we use in the experiments.

\FigureSamplerAndMixing

\subsection{Adaptive Location Sampling}
As discussed in Section~\ref{sec:revisiting}, the decoder should adaptively decide which feature to sample regarding the query.
That is, the decoder should decode sampling locations with the consideration of both the positional vector $(x, y, z, r)$ and content vector $\mathbf{q}$.
Also, we argue that the decoder must be adaptive not only over $(x,y)$ space but also be flexible in scales of potential objects. Specifically, we can accomplish these goals by regarding multi-scale features as a 3D feature space and adaptively sampling features from it.

\noindent\textbf{Multi-scale features as the 3D feature space.}
Given a feature map, indexed $j$, with the downsampling stride $s^\mathrm{feat}_j$ from the backbone, we first transform them by a linear layer to the same channel $d_{\rm feat}$ and compute its z-axis coordinate:
\begin{align}
    z^\mathrm{feat}_j = \log_{2}(s^\mathrm{feat}_j/s_{\rm base}).
\end{align}
Then we virtually rescale the height and width of feature maps of different strides to the same ones $H/s_{\rm base}$ and $W/s_{\rm base}$, where $H$ and $W$ is the height and width of the input image, and put them aligned on x- and y-axis in the 3D space as depicted in Figure~\ref{fig:sampler}. These feature maps are supporting planes for the 3D feature space, whose interpolation is described below.

\noindent\textbf{Adaptive 3D feature sampling process.}
A query first generate $P_{\rm in}$ sets of offset vectors to $P_{\rm in}$ points, $\{(\Delta x_i, \Delta y_i, \Delta z_i)\}_{P_{\rm in}}$, where each offset vector is indexed by $i$, and depends on its content vector $\mathbf{q}$ by a linear layer:
\begin{align}
    \{(\Delta x_i, \Delta y_i, \Delta z_i)\}_{P_{\rm in}} = {\rm Linear}(\mathbf{q}).
\end{align}
Then, these offsets are transformed to sampling locations according to the positional vector of the query for every $i$:
{\small
\begin{align}
    \begin{cases}
        \tilde{x}_i = x + \Delta x_i\cdot 2^{z-r}, \\
        \tilde{y}_i = y + \Delta y_i\cdot 2^{z+r}, \\
        \tilde{z}_i = z + \Delta z_i,
    \end{cases}
\end{align}
}

 It is worth noting that the area $\{\Delta x_{i},\Delta y_{i}\in[-0.5, 0.5]\}$ describes the bounding box decoded from the query. Our offsets are not restricted to this range, meaning that a query can sample features ``out of the box''. Having obtained $\{(\tilde{x}_i, \tilde{y}_i, \tilde{z}_i)\}_{P_{\rm in}}$, our sampler samples values given these points in the 3D space. In the current implementation, the interpolation over the 3D space is in the compositional manner: it first samples values given points by bilinear interpolation in the $(x, y)$ space and then interpolates over the z-axis by gaussian weighting given a sampling $\tilde{z}$, where the weight for the $j$-th feature map is: 
{\small
\begin{align}
    \tilde{w}_{j}=
    {\exp(-(\tilde{z} - z^\mathrm{feat}_j)^2/\tau_z) \over \sum_j \exp(-(\tilde{z} - z^\mathrm{feat}_j)^2/\tau_z)},
\end{align}
}
where $\tau_z$ is the softing coefficient for interpolating values over the z-axis and we keep $\tau_z=2$ in this work. With the feature map of the channel $d_{\rm feat}$, the shape of sampled feature matrix $\mathbf{x}$ is $\mathbb{R}^{P_{\rm in}\times d_{\rm feat}}$.
The adaptive 3D feature sampling process eases the decoder learning by sampling features with explicit, adaptive and coherent locations and scales regarding a query.

\noindent\textbf{Group sampling.}
To sample as many points as possible, we introduce the group sampling mechanism,  analogous to multiple heads in attentional operators~\cite{DBLP:conf/nips/VaswaniSPUJGKP17} or groups in group convolution~\cite{DBLP:conf/cvpr/XieGDTH17}. The group sampling first splits the channel $d_{\rm feat}$ of the 3D feature space into $g$ groups, each with the channel $d_{\rm feat}/g$, and performs 3D sampling individually for each group. With the group sampling mechanism, the decoder can generate $g\cdot P_{\rm in}$ offset vectors for a query to enrich the diversity of sampling points and exploit richer spatial structure of these points. Sampled feature matrix $\mathbf{x}$ now are of the shape $\mathbb{R}^{g\times P_{\rm in}\times (d_{\rm feat}/g)}$. The grouping mechanism is also applied to the adaptive mixing for efficiency as described below, and we term the group sampling and mixing unified as the grouping mechanism.

\FigureStructure

\subsection{Adaptive Content Decoding}
With features sampled, {\em how to adaptively decode them} is another key design in our AdaMixer decoder. To capture correlation in spatial and channel dimension of $\mathbf{x}$, we propose to efficiently decode the content in each dimension separately. Specifically, we design a simplified and adaptive variant of MLP-mixer~\cite{DBLP:journals/corr/abs-2105-01601}, termed as adaptive mixing, with dynamic mixing weights similar to dynamic filters in convolutions~\cite{DBLP:conf/nips/JiaBTG16}. As shown in Figure~\ref{fig:mixing}, the procedure contains sequentially the adaptive channel mixing and adaptive spatial mixing to involve both adaptive {\em channel semantics} and {\em spatial structures} under the guidance of a query.

\noindent\textbf{Adaptive channel mixing.}
Given sampled feature matrix $\mathbf{x}\in\mathbb{R}^{P_{\rm in}\times C}$ for a query in a group, where $C=d_{\rm feat}/g$, the adaptive channel mixing (ACM) is to use the dynamic weight based on $\mathbf{q}$ to transform features $\mathbf{x}$ on the channel dimension to adaptively enhance channel semantics:
\begin{align}
    M_c &= {\rm Linear}(\mathbf{q}) \in \mathbb{R}^{C\times C} \\
    {\rm ACM}(\mathbf{x})&= {\rm ReLU}({\rm LayerNorm}(\mathbf{x}M_c)),
\end{align}
where ${\rm ACM}(\mathbf{x})\in \mathbb{R}^{P_{\rm in}\times C}$ is the channel mixed feature output and the linear layer is individual for each group. The layer normalization~\cite{DBLP:journals/corr/BaKH16} is applied to both dimensions of the mixed output. Note that in this step, the dynamic weight is {\em shared} across {\em different sampling points} in 3D space,  analogous to adaptive $1\times 1$ convolution in~\cite{DBLP:journals/corr/abs-2011-12450} on RoI features.

\noindent\textbf{Adaptive spatial mixing.}
To enable the adaptability of a query to spatial structures of sampled features, we introduce the adaptive spatial mixing (ASM) process.
As depicted in Figure~\ref{fig:mixing}, ASM can be described as firstly transposing channel mixed feature matrix and applying the dynamic kernel to the spatial dimension of it:
\begin{align}
    M_s &= {\rm Linear}(\mathbf{q}) \in \mathbb{R}^{P_{\rm in}\times P_{\rm out}} \\
    {\rm ASM}(\mathbf{x}) &= {\rm ReLU}({\rm LayerNorm}(\mathbf{x}^T M_s)),
\end{align}
where ${\rm ASM}(\mathbf{x})\in \mathbb{R}^{C\times P_{\rm out}}$ is the spatial mixed output and $P_{\rm out}$ is the number of spatial mixing out patterns. Note that the dynamic weight is {\em shared} across {\em different channels}. As sampling points may be from different feature scales, ASM naturally involves multi-scale interaction modeling, which is necessary for high performance object detection.

The adaptive mixing procedure overall is depicted in Figure~\ref{fig:mixing}, where the adaptive spatial mixing follows the adaptive channel mixing, both applied in a sampling group. The final output of the shape $\mathbb{R}^{g\times C\times P_{\rm out}}$ across group is flattened and transformed to the $d_{q}$ dimension by a linear layer to add back to the content vector.

\subsection{Overall AdaMixer Detector}\label{sec:decoder}
Like the decoder architecture in~\cite{DBLP:conf/eccv/CarionMSUKZ20, DBLP:journals/corr/abs-2010-04159}, we place the self-attention between queries, our proposed adaptive mixing, and feedforward-feed network (FFN) sequentially in a stage of the decoder regarding the query content vector $\mathbf{q}$, as shown in Figure~\ref{fig:structure}. The query positional vector is updated by another FFN at the end of each stage:
{\small
\begin{align}
    &x' = x + \Delta x\cdot 2^z, y' = y + \Delta y\cdot 2^z,\\
    &z' = z + \Delta z, \hspace{1.9em} r' = r + \Delta r,
 \end{align}
}
where $(\Delta x, \Delta y, \Delta z, \Delta r)$ is produced by the lower small FFN block in Figure~\ref{fig:structure}.

\noindent\textbf{Position-aware multi-head self-attentions.}
Since we disentangle the content and position for a query, the naive multi-head self-attention between the content vectors of queries is not aware of {\em what geometric relation} between a query and another query is, which is proven beneficial to suppress redundant detections~\cite{DBLP:conf/eccv/CarionMSUKZ20}. To achieve this, we embed positional information into the self-attention. Our positional embedding for the content vector in the sinusoidal form and every component of $(x, y, z, r)$ takes up a quarter of channels.
We also embed the intersection over foreground (IoF) as a bias to the attention weight between queries to explicitly incorporate the relation of being contained between queries. The attention for each head is
{\small
\begin{align}
    {\rm Attn}(Q, K, V) = {\rm Softmax}\left(QK^T/\sqrt{d_{q}}+\alpha B\right)V,
\end{align}
}
where
$
    B_{ij} = \log\left({|{\rm box}_i\cap {\rm box}_j| / |{\rm box}_i}| + \epsilon\right)
$, $\epsilon=10^{-7}$, $Q, K, V\in\mathbb{R}^{N\times d_{q}}$, standing for the query, key and value matrix in the self-attention procedure, and $\alpha$ is a learnable scalar for each head. The $B_{ij}=0$ stands for the box $i$ being totally contained in the box $j$ and $B_{ij}=\log \epsilon \ll 0$ indicates that there is no overlapping between $i$ and $j$.

\noindent\textbf{Overall AdaMixer detector.}
The detection pipeline of AdaMixer is only composed of a backbone and a AdaMixer decoder. It avoids adding explicit feature pyramid networks or attentional encoders between backbone and decoder. The AdaMixer directly gathers predictions of decoded queries as final object detection results.

%% file: experiment.tex
\section{Experiments}
In this section, we first elaborate on the implementation and training details. Then we compare our models with other competitive detectors with limited training epochs. Next, we perform ablation studies on the design of our detector. We also align the training recipe to other query-based detectors and compare our AdaMixer to them fairly.

\subsection{Implementation Details}
\noindent\textbf{Dataset.}
We conduct extensive experiments on MS COCO 2017 dataset~\cite{DBLP:conf/eccv/LinMBHPRDZ14} in mmdetection codebase~\cite{mmdetection2018}. Following the common practice, we use \texttt{trainval35k} subset consisting up of 118K images to train our models and use \texttt{minival} subset of 5K images as the validation set.

\TableConfiguration

\noindent\textbf{Configurations.}
The default hyper-parameters in our AdaMixer detector is elaborated in Table~\ref{tab:configuration}.
We configure the dimension of the query content vector $d_q$ to $256$ following previous query-based work~\cite{DBLP:conf/eccv/CarionMSUKZ20, DBLP:journals/corr/abs-2010-04159, DBLP:journals/corr/abs-2011-12450}. We use feature maps $C_2\sim C_5$ from the backbone network. Multi-scale features are processed by the linear transformation to the channel $d_{\rm feat}=256$ as supporting planes for the 3D feature space. The number of decoder stages is set to 6 also following the common practice of query-based detectors. The mixer grouping number is set to 4 as default. Accordingly, the channel of sampled features per group is $C=d_{\rm feat}/g=64$. Also, the number of sampling points $P_{\rm in}$ and mixing out patterns $P_{\rm out}$ per group is set to $32$ and $128$. The hidden dimension of FFN on the content vector in the decoder is set to 2048. 
The FFN dimension for classification and updating positional vectors is set to 256.

\noindent\textbf{Initializations.}
For training stability in early iterations, we initialize the parameters of linear layers to produce dynamic parameters or sampling offsets as follows: zeroing weights of these layers and initializing biases as expected. This helps stable training by enforcing models to learn the adaptability from zeros. The biases for the linear layer producing sampling offest vectors is initialized in the way that $\Delta x_{i}, \Delta y_{i}$ are uniformly drawn in $[-0.5, 0.5]$ and $\Delta z_{i}=-1$ for all $i$ to align with the RoIAlign~\cite{DBLP:conf/iccv/HeGDG17} level strategy. The bias in the linear layer to produce mixing weights follows the default initialization in the PyTorch. We also initialize all the query positional vectors into decoders such that the their boxes and sampling points cover the whole image in the initial decoder stage like~\cite{DBLP:journals/corr/abs-2011-12450}.
Backbones are initialized from pre-trained models on the ImageNet 1K dataset~\cite{DBLP:conf/cvpr/DengDSLL009}.

\TableOneTimesSchedule

\noindent\textbf{Losses and optimizers.}
Following~\cite{DBLP:journals/corr/abs-2010-04159, DBLP:journals/corr/abs-2011-12450}, the training loss is the matching loss consisting of the focal loss~\cite{DBLP:conf/iccv/LinGGHD17} with coefficient $\lambda_{\rm cls}=2$, L1 bounding box loss with $\lambda_{L_1}=5$ and GIoU loss~\cite{DBLP:conf/cvpr/RezatofighiTGS019} with $\lambda_{\rm giou}=2$.
We use AdamW~\cite{DBLP:conf/iclr/LoshchilovH19} as our optimizer with weight decay 0.0001. The initial learning rate is $2.5\times 10^{-5}$.

\TableAblationStudiesIntegrated

\noindent\textbf{Training recipes.}
We adopt two versions of training recipes for a fair comparison with different detectors. The first one adopts the classic \textbf{$\mathbf{1\times} $ training scheme}, which includes a budget of 12 training epochs with training images of shorter side resized to 800. This recipe includes the random horizontal flipping as only the standard data augmentation and allocates 100 learnable object queries to our AdaMixer detector, to compare with popular and competitive detectors like FCOS~\cite{DBLP:conf/iccv/TianSCH19} and Cascade R-CNN~\cite{DBLP:conf/cvpr/CaiV18} fairly. The second training recipe is to align with other query-based detectors, which leverages more training epochs and performs crop and multi-scale augmentation in~\cite{DBLP:conf/eccv/CarionMSUKZ20, DBLP:journals/corr/abs-2010-04159, DBLP:journals/corr/abs-2011-12450}. Our second training recipe adopts the same data augmentation and has a budget of 36 training epochs, namely \textbf{$\mathbf{3\times}$ training scheme}. It uses 300 object queries to compare fairly with~\cite{DBLP:journals/corr/abs-2010-04159, DBLP:journals/corr/abs-2011-12450}. The learning rate is divided by a factor of 10 at epoch 8 and 11 in $1\times$ training scheme or at epoch 24 and 33 in $3\times$ training scheme, scaled proportionally.

The $1\times$ and $3\times$ training scheme for AdaMixer with ResNet-50 typically take about 9 and 29 hours on 8 V100 cards.
During the inference stage, we input images of the shorter size resized to 800 without data augmentation. We leave more details about model training and inference and visualizations in the supplementary material.
\subsection{Fast Convergence with Limited Budgets}
We first investigate our proposed AdaMixer with limited training epochs and limited data augmentation, namely $1\times$ training scheme. For a fair comparison, we disable the commonly-used crop and multi-scale data augmentation in query-based detectors and allocate only 100 queries or learnable proposals for these detectors. 
Experimental results are shown in Table~\ref{table:onetimeschedule}.
AdaMixer with $N=100$ queries achieves 42.7 AP, outperforming state-of-the-art traditional and query-based detectors with a limited training budget.
Moreover, if we increase the number of queries $N$ to 300 and 500, the performance of the AdaMixer detector reaches 44.1 and 45.0 AP, especially with 27.0 and 27.9 AP$_s$ in detecting small objects. It is worth noting that these results are achieved with random flipping as the only data augmentation and within 12 training epochs, showing that AdaMixer can be supervised efficiently with training samples.

\subsection{Ablation Studies}
Due to the limited computational resource, we use ResNet-50 as the backbone network and $1\times$ training scheme to perform ablation studies.

\noindent\textbf{Decoding adaptability.}
We begin our ablations with the key component, the adaptability design, in our AdaMixer detector. The adaptability in AdaMixer is in two aspects: adaptive sampling for decoding locations and adaptive mixing for decoding content. Table~\ref{tab:adaptabilityDecoding} investigates the performance under the condition of whether or not we enable the adaptability in decoding sampling locations and decoding content. The cancellation for adaptability on locations or content stands for enforcing weights of linear layers producing sampling offsets or mixing weights all to zeros during the training and inference. Only biases of these layers can be learned during the training procedure, which are eventually not adaptive based on the query content $\mathbf{q}$. In other words, all sampling offsets or mixing weights are the same across different queries and different images with the cancellation. As shown in Table~\ref{tab:adaptabilityDecoding}, the adaptability in both decoding sampling locations and sampled content is essential to a good query-based object detector, which outperforms a non-adaptive counterpart by 7.0 AP.

\noindent\textbf{Adaptive mixing design.}
Moving forward, we compare different designs of our adaptive mixing in Table~\ref{tab:mixingDesign}. As shown in Figure~\ref{fig:mixing}, our default design for adaptive mixing is to mix features first on the channel dimension and then on the spatial dimension. We perform ablations by placing only channel mixing, only spatial mixing, and the reversed order of our design as three variants. The first adaptive channel mixing and then spatial one lead to the best performance. This indicates that channel semantics and spatial structures are both important to the mixing design. For the reversed mixing variant, we suspect that the inferior result is due to the insufficient channel semantics into spatial mixing as features are directly from the backbone.

\TableSOTA

\noindent\textbf{Extra pyramid networks.}
AdaMixer enjoys the simplicity for circumventing extra attentional encoders or explicit pyramid networks. Instead, AdaMixer improves the semantic and multi-scale modeling in the decoder. The adaptive 3D sampling and following spatial mixing naturally enable multi-scale feature modeling and enable queries to handle scale variations of objects. In Table~\ref{tab:extraNeck}, we investigate the performance of the AdaMixer detector with introduction of extra pyramid networks. Models with these extra networks might require a longer training time and more training samples to perform well. These results are in favor of our AdaMixer design as a simplified query-based detector.

\noindent\textbf{Sampling points and spatial mixing out patterns.}
Table~\ref{tab:pin} and \ref{tab:pout} shows the ablation on the sampling points $P_{\rm in}$ and spatial mixing out patterns $P_{\rm out}$ per group. The performance is generally related to the number of sampling points $P_{\rm in}$ and spatial mixing out patterns $P_{\rm out}$.
A good balance between the complexity and performance is $P_{\rm in}=32$ and $P_{\rm out}=128$, where the performance saturates for $P_{\rm in}$ and decreases for $P_{\rm out}$ beyond this point.

\noindent\textbf{Positional information in attention between queries.}
In Section~\ref{sec:decoder}, we propose to embed the positional information into the self-attention between the content vectors of queries.
In addition to the regular sinusoidal positional embedding, we also hardwire the intersection over foreground (IoF) into the attention weight between boxes indicated by queries.
We investigate these two ingredients in Table~\ref{tab:positionalInformation}. Results show that combining these two ingredients notably increases the performance. The individual effect of the IoF is also compelling. We argue that the IoF between boxes, which describes the geometric relation of being contained directly for corresponding queries, is important for the self-attention to imitate the NMS procedure~\cite{DBLP:conf/eccv/CarionMSUKZ20}.

\TableGroup
\noindent\textbf{Mixer group number.}
The mixer grouping encourages the decoder sampler to sample more diverse points. This also reduces total parameters and computational costs by mixing divided groups of features. We here evaluate the effect of the grouping mechanism with various $g$ in Table~\ref{tab:group}. The model reaches the least FLOPs and number of parameters with $4$ mixer groups with promising performance. 

\subsection{Comparison with Other Query-based Detectors}
We present the final results of our AdaMixer and perform the comparison between our AdaMixer and other state-of-the-art query-based detectors in Table~\ref{tab:longscheme}.
We use the $3\times$ training scheme to train our AdaMixer, which allocates 300 queries and includes the stronger data augmentation to align with the common practice of other query-based methods.
Specifically, we train our AdaMixer with ResNet-50, ResNet-101, ResNeXt-101~\cite{DBLP:conf/cvpr/XieGDTH17} with deformable convolution layers~\cite{DBLP:journals/corr/abs-1811-11168} and Swin-S~\cite{DBLP:journals/corr/abs-2103-14030} backbones.
We also proportionally stretch the training schedule for AdaMixer with ResNet-50 to investigate the faster convergence speed, as depicted in Figure~\ref{fig:curve}. 
AdaMixer with assorted backbones significantly outperforms competitive query-based object detectors with less computational cost.
With bounding boxes as only supervising signals, AdaMixer with Swin-S reaches 51.3 AP and 34.2 AP$_s$ with the single scale testing.
Moreover, among these query-based detectors, only AdaMixer does not require the extra attentional encoders and explicit pyramid networks. 
These results demonstrate our AdaMixer is a simply-architected, effective, and fast-converging query-based object detector.

%% file: supp.tex
\section*{Appendix}
\section*{A. 1. Detection Performance on COCO test set}
\TableTestDev
The performance of AdaMixer on COCO~\texttt{minival} set is reported in Table~\ref{tab:longscheme} in the paper. Here we report the performance of these AdaMixer models on COCO~\texttt{test-dev} set in Table~\ref{tab:testdev}, where labels are not publicly available and evaluation is done on the online server. Note that models here are exactly the same in Table~5 in the paper.

\section*{A. 2. More Studies on Adaptive 3D Sampling}
\TableAblationThreeDSampling
We here also validate the effectiveness of our proposed 3D feature space and adaptive 3D sampling in Table~\ref{tab:sampling}.
Using single feature map leads to inferior results for our AdaMixer.
Note that there is no feature pyramid networks (FPN)~\cite{DBLP:conf/cvpr/LinDGHHB17} used. The RoIAlign operator, which extracts features in a single level map according to the bounding box, also performs inferior to our adaptive 3D sampling approach. RoIAlign feature sampling locations are highly stricted inner the bounding box and the sampling level (the $z$ coordinate in our work) is also discretized, whereas locations and levels are adaptive on queries and not stricted in our adaptive 3D sampling . This means that RoIAlign lacks multi-level feature modeling and necessitates multi-scale feature interaction necks.

\noindent\textbf{Relations to other work.}
Our adaptive 3D sampling is in accord with~ deformable convolutions\cite{DBLP:conf/iccv/DaiQXLZHW17, DBLP:journals/corr/abs-1811-11168, DBLP:journals/corr/abs-2010-04159} to explicitly utilize offsets to sample features spatially to model deformation of objects. But our adaptive 3D sampling extends the adaptive modeling to the scale dimension, namely z-axis, to cope with the variation of potential objects. Also, our adaptive 3D sampling is performed by queries in the sparse manner, whereas the deformable convolution or attention are usually performed in the dense manner. 
Moreover, our method naturally generalizes the scale-equalizing operators~\cite{DBLP:journals/corr/abs-1901-01892,DBLP:conf/cvpr/WangZYFZ20} on discretized feature maps with different scales to a continuous and interpolable one, enjoying more flexible modeling for scale variations.

\section*{A. 3. Model Details}
Our AdaMixer is implemented on mmdetection framework~\cite{mmdetection2018}. In current implementation, the model is fully based on PyTorch primitives without customized CUDA codes.

\TableModel
\noindent\textbf{Initializations.}
We initialize all backbones from pre-trained model on ImageNet \textbf{1K} classification~\cite{DBLP:conf/cvpr/DengDSLL009}, including Swin-S~\cite{DBLP:journals/corr/abs-2103-14030}. Please refer to our codebase for more detailed initializations in AdaMixer.

\noindent\textbf{Training and inference speed.}
We evaluate AdaMixer training and inference speed and compare it with other detectors in Table~\ref{tab:model}. Our AdaMixer shows advantage not only on theoretical flops but on the actual training time and inference FPS.
Note that the current implementation, the adaptive 3D sampling procedure is based on PyTorch primitive \texttt{grid\_sample}, which can be optimized in the future.

\section*{A. 4. Visualizations}
We also show visualizations of sampling points and final detections of our AdaMixer detector in Figure~\ref{fig:a} and \ref{fig:b}.
These visualizations are doned on samples in COCO \texttt{minival} set.
The first row shows the input image and detection results and the other rows show sampling points for each stage. Figure~\ref{fig:a} and \ref{fig:b} shows two different query results\footnote{In a figure, the query index is consistent through stages.}. The sampling z-axis coordinate, $\tilde{z}$, is visualized as the point size and a bigger point corresponds with the larger $\tilde{z}$.
Sampling points of different groups are colored with different colors.
We can see that AdaMixer actually sees out of the box and different groups have preference for different semantics. Moreover, when comparing Figure~\ref{fig:a} and \ref{fig:b}, we can find that sampling point patterns also vary across different queries, indicating the enhanced sampling adaptability and flexibility in AdaMixer.  Counterintuitively, sampling points of a query do not gather together in a monotonous manner across stages. Instead, they often stretch wider after focusing on small regions and then focus again (stage 2$\to$3$\to$4$\to$5$\to$6). We suspect that this behavior is beneficial to performing bounding box estimation more accurately. Please see \texttt{visualizations} folder in our codebase for more visualizations.

\FigSuppA
\FigSuppB

%% file: cvpr.bbl
\begin{thebibliography}{10}\itemsep=-1pt

\bibitem{DBLP:journals/corr/BaKH16}
Lei~Jimmy Ba, Jamie~Ryan Kiros, and Geoffrey~E. Hinton.
\newblock Layer normalization.
\newblock {\em arXiv}, 2016.

\bibitem{DBLP:conf/iccv/BodlaSCD17}
Navaneeth Bodla, Bharat Singh, Rama Chellappa, and Larry~S. Davis.
\newblock Soft-nms - improving object detection with one line of code.
\newblock In {\em {ICCV}}, 2017.

\bibitem{DBLP:conf/cvpr/CaiV18}
Zhaowei Cai and Nuno Vasconcelos.
\newblock Cascade {R-CNN:} delving into high quality object detection.
\newblock In {\em {CVPR}}, 2018.

\bibitem{DBLP:conf/eccv/CarionMSUKZ20}
Nicolas Carion, Francisco Massa, Gabriel Synnaeve, Nicolas Usunier, Alexander
  Kirillov, and Sergey Zagoruyko.
\newblock End-to-end object detection with transformers.
\newblock In {\em ECCV}, 2020.

\bibitem{DBLP:conf/cvpr/ChenPWXLSF0SOLL19}
Kai Chen, Jiangmiao Pang, Jiaqi Wang, Yu Xiong, Xiaoxiao Li, Shuyang Sun,
  Wansen Feng, Ziwei Liu, Jianping Shi, Wanli Ouyang, Chen~Change Loy, and
  Dahua Lin.
\newblock Hybrid task cascade for instance segmentation.
\newblock In {\em {CVPR}}, 2019.

\bibitem{mmdetection2018}
Kai Chen, Jiaqi Wang, Jiangmiao Pang, Yuhang Cao, Yu Xiong, Xiaoxiao Li,
  Shuyang Sun, Wansen Feng, Ziwei Liu, Jiarui Xu, Zheng Zhang, Dazhi Cheng,
  Chenchen Zhu, Tianheng Cheng, Qijie Zhao, Buyu Li, Xin Lu, Rui Zhu, Yue Wu,
  Jifeng Dai, Jingdong Wang, Jianping Shi, Wanli Ouyang, Chen~Change Loy, and
  Dahua Lin.
\newblock Mmdetection: Open mmlab detection toolbox and benchmark.
\newblock In {\em arXiv}, 2019.

\bibitem{DBLP:conf/iccv/DaiQXLZHW17}
Jifeng Dai, Haozhi Qi, Yuwen Xiong, Yi Li, Guodong Zhang, Han Hu, and Yichen
  Wei.
\newblock Deformable convolutional networks.
\newblock In {\em ICCV}, 2017.

\bibitem{DBLP:conf/cvpr/DaiCX0LY021}
Xiyang Dai, Yinpeng Chen, Bin Xiao, Dongdong Chen, Mengchen Liu, Lu Yuan, and
  Lei Zhang.
\newblock Dynamic head: Unifying object detection heads with attentions.
\newblock In {\em {CVPR}}, 2021.

\bibitem{DBLP:conf/cvpr/DengDSLL009}
Jia Deng, Wei Dong, Richard Socher, Li{-}Jia Li, Kai Li, and Fei{-}Fei Li.
\newblock Imagenet: {A} large-scale hierarchical image database.
\newblock In {\em CVPR}, 2009.

\bibitem{DBLP:conf/iccv/DuanBXQH019}
Kaiwen Duan, Song Bai, Lingxi Xie, Honggang Qi, Qingming Huang, and Qi Tian.
\newblock Centernet: Keypoint triplets for object detection.
\newblock In {\em {ICCV}}, 2019.

\bibitem{DBLP:conf/cvpr/FelzenszwalbGM10}
Pedro~F. Felzenszwalb, Ross~B. Girshick, and David~A. McAllester.
\newblock Cascade object detection with deformable part models.
\newblock In {\em {CVPR}}, 2010.

\bibitem{DBLP:journals/pami/FelzenszwalbGMR10}
Pedro~F. Felzenszwalb, Ross~B. Girshick, David~A. McAllester, and Deva Ramanan.
\newblock Object detection with discriminatively trained part-based models.
\newblock {\em {IEEE} Trans. Pattern Anal. Mach. Intell.}, 32(9):1627--1645,
  2010.

\bibitem{DBLP:journals/corr/abs-2101-07448}
Peng Gao, Minghang Zheng, Xiaogang Wang, Jifeng Dai, and Hongsheng Li.
\newblock Fast convergence of {DETR} with spatially modulated co-attention.
\newblock In {\em ICCV}, 2021.

\bibitem{DBLP:conf/iccv/Gao0W21}
Ziteng Gao, Limin Wang, and Gangshan Wu.
\newblock Mutual supervision for dense object detection.
\newblock In {\em ICCV}, 2021.

\bibitem{DBLP:conf/cvpr/GeLLYS21}
Zheng Ge, Songtao Liu, Zeming Li, Osamu Yoshie, and Jian Sun.
\newblock {OTA:} optimal transport assignment for object detection.
\newblock In {\em {CVPR}}, 2021.

\bibitem{DBLP:conf/iccv/Girshick15}
Ross~B. Girshick.
\newblock Fast {R-CNN}.
\newblock In {\em ICCV}, 2015.

\bibitem{DBLP:conf/iccv/HeGDG17}
Kaiming He, Georgia Gkioxari, Piotr Doll{\'{a}}r, and Ross~B. Girshick.
\newblock Mask {R-CNN}.
\newblock In {\em ICCV}, 2017.

\bibitem{DBLP:conf/cvpr/HeZRS16}
Kaiming He, Xiangyu Zhang, Shaoqing Ren, and Jian Sun.
\newblock Deep residual learning for image recognition.
\newblock In {\em CVPR}, 2016.

\bibitem{DBLP:conf/cvpr/HeZWS019}
Yihui He, Chenchen Zhu, Jianren Wang, Marios Savvides, and Xiangyu Zhang.
\newblock Bounding box regression with uncertainty for accurate object
  detection.
\newblock In {\em {CVPR}}, 2019.

\bibitem{DBLP:conf/nips/JiaBTG16}
Xu Jia, Bert~De Brabandere, Tinne Tuytelaars, and Luc~Van Gool.
\newblock Dynamic filter networks.
\newblock In {\em {NIPS}}, 2016.

\bibitem{DBLP:conf/eccv/LawD18}
Hei Law and Jia Deng.
\newblock Cornernet: Detecting objects as paired keypoints.
\newblock In {\em {ECCV}}, 2018.

\bibitem{DBLP:conf/cvpr/LiW0LT021}
Xiang Li, Wenhai Wang, Xiaolin Hu, Jun Li, Jinhui Tang, and Jian Yang.
\newblock Generalized focal loss {V2:} learning reliable localization quality
  estimation for dense object detection.
\newblock In {\em {CVPR}}, 2021.

\bibitem{DBLP:journals/corr/abs-1901-01892}
Yanghao Li, Yuntao Chen, Naiyan Wang, and Zhaoxiang Zhang.
\newblock Scale-aware trident networks for object detection.
\newblock In {\em arXiv}, 2019.

\bibitem{DBLP:conf/cvpr/LinDGHHB17}
Tsung{-}Yi Lin, Piotr Doll{\'{a}}r, Ross~B. Girshick, Kaiming He, Bharath
  Hariharan, and Serge~J. Belongie.
\newblock Feature pyramid networks for object detection.
\newblock In {\em {CVPR}}, 2017.

\bibitem{DBLP:conf/iccv/LinGGHD17}
Tsung{-}Yi Lin, Priya Goyal, Ross~B. Girshick, Kaiming He, and Piotr
  Doll{\'{a}}r.
\newblock Focal loss for dense object detection.
\newblock In {\em {ICCV}}, 2017.

\bibitem{DBLP:conf/eccv/LinMBHPRDZ14}
Tsung{-}Yi Lin, Michael Maire, Serge~J. Belongie, James Hays, Pietro Perona,
  Deva Ramanan, Piotr Doll{\'{a}}r, and C.~Lawrence Zitnick.
\newblock Microsoft {COCO:} common objects in context.
\newblock In {\em ECCV}, 2014.

\bibitem{DBLP:conf/cvpr/LiuQQSJ18}
Shu Liu, Lu Qi, Haifang Qin, Jianping Shi, and Jiaya Jia.
\newblock Path aggregation network for instance segmentation.
\newblock In {\em {CVPR}}, 2018.

\bibitem{DBLP:conf/eccv/LiuAESRFB16}
Wei Liu, Dragomir Anguelov, Dumitru Erhan, Christian Szegedy, Scott~E. Reed,
  Cheng{-}Yang Fu, and Alexander~C. Berg.
\newblock {SSD:} single shot multibox detector.
\newblock In {\em {ECCV}}, 2016.

\bibitem{DBLP:journals/corr/abs-2103-14030}
Ze Liu, Yutong Lin, Yue Cao, Han Hu, Yixuan Wei, Zheng Zhang, Stephen Lin, and
  Baining Guo.
\newblock Swin transformer: Hierarchical vision transformer using shifted
  windows.
\newblock In {\em {ICCV}}, 2021.

\bibitem{DBLP:conf/iclr/LoshchilovH19}
Ilya Loshchilov and Frank Hutter.
\newblock Decoupled weight decay regularization.
\newblock In {\em {ICLR}}, 2019.

\bibitem{DBLP:journals/corr/abs-2108-06152}
Depu Meng, Xiaokang Chen, Zejia Fan, Gang Zeng, Houqiang Li, Yuhui Yuan, Lei
  Sun, and Jingdong Wang.
\newblock Conditional {DETR} for fast training convergence.
\newblock In {\em ICCV}, 2021.

\bibitem{DBLP:conf/cvpr/PangCSFOL19}
Jiangmiao Pang, Kai Chen, Jianping Shi, Huajun Feng, Wanli Ouyang, and Dahua
  Lin.
\newblock Libra {R-CNN:} towards balanced learning for object detection.
\newblock In {\em {CVPR}}, 2019.

\bibitem{DBLP:conf/eccv/QiuMLLS20}
Han Qiu, Yuchen Ma, Zeming Li, Songtao Liu, and Jian Sun.
\newblock Borderdet: Border feature for dense object detection.
\newblock In {\em {ECCV}}, 2020.

\bibitem{DBLP:conf/cvpr/RedmonDGF16}
Joseph Redmon, Santosh~Kumar Divvala, Ross~B. Girshick, and Ali Farhadi.
\newblock You only look once: Unified, real-time object detection.
\newblock In {\em {CVPR}}, 2016.

\bibitem{DBLP:journals/corr/abs-1804-02767}
Joseph Redmon and Ali Farhadi.
\newblock Yolov3: An incremental improvement.
\newblock {\em arXiv}, 2018.

\bibitem{DBLP:conf/nips/RenHGS15}
Shaoqing Ren, Kaiming He, Ross~B. Girshick, and Jian Sun.
\newblock Faster {R-CNN:} towards real-time object detection with region
  proposal networks.
\newblock In {\em {NIPS}}, 2015.

\bibitem{DBLP:conf/cvpr/RezatofighiTGS019}
Hamid Rezatofighi, Nathan Tsoi, JunYoung Gwak, Amir Sadeghian, Ian~D. Reid, and
  Silvio Savarese.
\newblock Generalized intersection over union: {A} metric and a loss for
  bounding box regression.
\newblock In {\em {CVPR}}, 2019.

\bibitem{DBLP:journals/corr/SermanetEZMFL13}
Pierre Sermanet, David Eigen, Xiang Zhang, Micha{\"{e}}l Mathieu, Rob Fergus,
  and Yann LeCun.
\newblock Overfeat: Integrated recognition, localization and detection using
  convolutional networks.
\newblock In {\em ICLR}, 2014.

\bibitem{DBLP:journals/corr/abs-2011-12450}
Peize Sun, Rufeng Zhang, Yi Jiang, Tao Kong, Chenfeng Xu, Wei Zhan, Masayoshi
  Tomizuka, Lei Li, Zehuan Yuan, Changhu Wang, and Ping Luo.
\newblock Sparse {R-CNN:} end-to-end object detection with learnable proposals.
\newblock In {\em CVPR}, 2021.

\bibitem{DBLP:conf/iccv/TianSCH19}
Zhi Tian, Chunhua Shen, Hao Chen, and Tong He.
\newblock {FCOS:} fully convolutional one-stage object detection.
\newblock In {\em {ICCV}}, 2019.

\bibitem{DBLP:journals/corr/abs-2105-01601}
Ilya~O. Tolstikhin, Neil Houlsby, Alexander Kolesnikov, Lucas Beyer, Xiaohua
  Zhai, Thomas Unterthiner, Jessica Yung, Andreas Steiner, Daniel Keysers,
  Jakob Uszkoreit, Mario Lucic, and Alexey Dosovitskiy.
\newblock Mlp-mixer: An all-mlp architecture for vision.
\newblock In {\em NeurIPS}, 2021.

\bibitem{DBLP:conf/nips/VaswaniSPUJGKP17}
Ashish Vaswani, Noam Shazeer, Niki Parmar, Jakob Uszkoreit, Llion Jones,
  Aidan~N. Gomez, Lukasz Kaiser, and Illia Polosukhin.
\newblock Attention is all you need.
\newblock In {\em {NIPS}}, 2017.

\bibitem{DBLP:conf/cvpr/ViolaJ01}
Paul~A. Viola and Michael~J. Jones.
\newblock Rapid object detection using a boosted cascade of simple features.
\newblock In {\em {CVPR}}, 2001.

\bibitem{DBLP:conf/cvpr/WangCYLL19}
Jiaqi Wang, Kai Chen, Shuo Yang, Chen~Change Loy, and Dahua Lin.
\newblock Region proposal by guided anchoring.
\newblock In {\em {CVPR}}, 2019.

\bibitem{DBLP:conf/cvpr/WangZYFZ20}
Xinjiang Wang, Shilong Zhang, Zhuoran Yu, Litong Feng, and Wayne Zhang.
\newblock Scale-equalizing pyramid convolution for object detection.
\newblock In {\em {CVPR}}, 2020.

\bibitem{DBLP:journals/corr/abs-2109-07107}
Yingming Wang, Xiangyu Zhang, Tong Yang, and Jian Sun.
\newblock Anchor {DETR:} query design for transformer-based detector.
\newblock {\em arXiv}, 2021.

\bibitem{DBLP:conf/cvpr/XieGDTH17}
Saining Xie, Ross~B. Girshick, Piotr Doll{\'{a}}r, Zhuowen Tu, and Kaiming He.
\newblock Aggregated residual transformations for deep neural networks.
\newblock In {\em {CVPR}}, 2017.

\bibitem{DBLP:conf/nips/YangZLZS18}
Tong Yang, Xiangyu Zhang, Zeming Li, Wenqiang Zhang, and Jian Sun.
\newblock Metaanchor: Learning to detect objects with customized anchors.
\newblock In {\em NeurIPS}, 2018.

\bibitem{DBLP:journals/corr/abs-2104-01318}
Zhuyu Yao, Jiangbo Ai, Boxun Li, and Chi Zhang.
\newblock Efficient {DETR:} improving end-to-end object detector with dense
  prior.
\newblock {\em ArXiv}, 2021.

\bibitem{DBLP:conf/mm/YuJWCH16}
Jiahui Yu, Yuning Jiang, Zhangyang Wang, Zhimin Cao, and Thomas~S. Huang.
\newblock Unitbox: An advanced object detection network.
\newblock In {\em {ACM} Multimedia}, 2016.

\bibitem{DBLP:journals/corr/abs-1912-02424}
Shifeng Zhang, Cheng Chi, Yongqiang Yao, Zhen Lei, and Stan~Z. Li.
\newblock Bridging the gap between anchor-based and anchor-free detection via
  adaptive training sample selection.
\newblock In {\em CVPR}, 2020.

\bibitem{DBLP:conf/nips/ZhangWLJY19}
Xiaosong Zhang, Fang Wan, Chang Liu, Rongrong Ji, and Qixiang Ye.
\newblock Freeanchor: Learning to match anchors for visual object detection.
\newblock In {\em NeurIPS}, 2019.

\bibitem{DBLP:journals/corr/abs-1904-07850}
Xingyi Zhou, Dequan Wang, and Philipp Kr{\"{a}}henb{\"{u}}hl.
\newblock Objects as points.
\newblock {\em ArXiv}, 2019.

\bibitem{DBLP:conf/cvpr/ZhouZK19}
Xingyi Zhou, Jiacheng Zhuo, and Philipp Kr{\"{a}}henb{\"{u}}hl.
\newblock Bottom-up object detection by grouping extreme and center points.
\newblock In {\em {CVPR}}, 2019.

\bibitem{DBLP:journals/corr/abs-1811-11168}
Xizhou Zhu, Han Hu, Stephen Lin, and Jifeng Dai.
\newblock Deformable convnets v2: More deformable, better results.
\newblock In {\em CVPR}, 2019.

\bibitem{DBLP:journals/corr/abs-2010-04159}
Xizhou Zhu, Weijie Su, Lewei Lu, Bin Li, Xiaogang Wang, and Jifeng Dai.
\newblock Deformable {DETR:} deformable transformers for end-to-end object
  detection.
\newblock In {\em ICLR}, 2020.

\end{thebibliography}
